\documentclass[12pt]{iopart}
\usepackage{iopams}  
\usepackage{gensymb}
\usepackage{hyperref}
\usepackage{listings}%
\usepackage{booktabs}
\usepackage{multirow}
\usepackage{graphicx}
\usepackage{xcolor}

\newcommand{\ket}[1]{|#1\rangle}
\newcommand{\bra}[1]{\langle #1|}

\begin{document}

\title[Benchmarking QNNs for Wind Energy Forecasting]{Quantum Neural Networks for Wind Energy Forecasting: A Comparative Study of Performance and Scalability with Classical Models}

\author{Batuhan Hangun$^{1,*}$, Oguz Altun$^1$, Onder Eyecioglu$^{2,\dagger}$}

\address{$^1$Department of Computer Engineering, Yildiz Technical University, Esenler, Istanbul, Türkiye}
\address{$^2$Department of Computer Engineering, Bolu Abant Izzet Baysal University, Golkoy, Bolu, Türkiye}

\eads{\mailto{batuhanhangun@gmail.com}, \mailto{onder.eyecioglu@ibu.edu.tr}}
\begin{indented}
\item[]$^*$Corresponding author: \mailto{batuhanhangun@gmail.com}
\item[]$^\dagger$Corresponding author: \mailto{onder.eyecioglu@ibu.edu.tr}
  \item[]ORCID: Batuhan Hangun — \href{https://orcid.org/0000-0002-0271-6868}{0000-0002-0271-6868}
  \item[]ORCID: Onder Eyecioglu — \href{https://orcid.org/0000-0002-9735-5697}{0000-0002-9735-5697}
\end{indented}

\begin{abstract}
Quantum Neural Networks (QNNs), a prominent approach in Quantum Machine Learning (QML), are emerging as a powerful alternative to classical machine learning methods. Recent studies have focused on the applicability of QNNs to various tasks, such as time-series forecasting, prediction, and classification, across a wide range of applications, including cybersecurity and medical imaging. With the increased use of smart grids driven by the integration of renewable energy systems, machine learning plays an important role in predicting power demand and detecting system disturbances. This study provides an in-depth investigation of QNNs for predicting the power output of a wind turbine. We assess the predictive performance and simulation time of six QNN configurations that are based on the Z Feature Map for data encoding and varying ansatz structures. Through detailed cross-validation experiments and tests on an unseen hold-out dataset, we experimentally demonstrate that QNNs can achieve predictive performance that is competitive with, and in some cases marginally better than, the benchmarked classical approaches. Our results also reveal the effects of dataset size and circuit complexity on predictive performance and simulation time. We believe our findings will offer valuable insights for researchers in the energy domain who wish to incorporate quantum machine learning into their work.
\end{abstract}

\vspace{2pc}
\noindent{\it Keywords}: quantum neural network, quantum circuit benchmark, simulation time complexity, quantum computing, wind power prediction

\section{Introduction}
Quantum computing has gained significant importance in recent years, largely due to advancements in quantum hardware~\cite{willow2024, majorona2025}. Despite the limitations of the noisy intermediate-scale quantum (NISQ) era, recent studies, such as those by Microsoft, indicate that large-scale quantum computing may still be years away~\cite{TomSvore2024, microsoft2024}. Nevertheless, such studies also build optimism that practically useful, large-scale quantum computing may be closer than previously expected. However, researchers striving to leverage the quantum advantage in fields currently dominated by classical computing remain cautious, emphasizing the need for more practical and demonstrable use cases to overcome skepticism. Therefore, detailed experimental studies that demonstrate the potential advantages of quantum computing are essential. In parallel with these hardware developments, another promising direction lies in the integration of quantum computing with artificial intelligence (AI).

As artificial intelligence (AI) becomes increasingly dominant in industry and academia, its integration with quantum computing represents a significant opportunity for achieving quantum advantage. Quantum machine learning (QML) aims to blend classical machine learning (ML) methods with quantum computing to develop hybrid models that can surpass their classical counterparts in learning and generalizing data. However, despite growing interest, many studies lack comprehensive benchmarking of quantum neural network (QNN) configurations, especially with respect to classical baselines on practical datasets. Since QML is still an emerging field, experimental studies are needed to systematically evaluate different quantum circuit designs (ansätze) to guide future model development and deployment.

To date, various studies focus on showing practical uses of QML to demonstrate its advantages over classical ML in different use cases. Researchers aim to either enhance classical ML methods using quantum counterparts or replace specific components with quantum analogs, forming hybrid models that can outperform pure classical ML methods. An application of unsupervised QML based on support vector machines by Kyriienko and Magnusson was deployed to detect credit card fraud. Their use of quantum kernels on fraud detection outperformed the classical kernel-based approach, achieving an additional 15\% precision gain, thus demonstrating potential advantages~\cite{kyriienko2022}. Sakhnenko et al. 
used a hybrid classical-quantum autoencoder (HAE) model, combining a classical autoencoder (AE) with a parametrized quantum circuit (PQC), to detect anomalies in three different datasets~\cite{sakhnenko2022hybrid}. In a similar anomaly detection study, Amin et al. proposed a quantum convolutional neural network (QCNN) approach for detecting anomalies in video streams. By using a hybrid model that combines classical CNN with quantum computing, they achieved 97\% accuracy~\cite{javaria2023}. Similarly, Wang et al. proposed a hybrid quantum-classical DNN (QHDNN) model that detects anomalies in images. Their experiments on MNIST and FashionMNIST show that the proposed hybrid model achieves up to 89\% detection accuracy and can outperform conventional DNN structures~\cite{wang2023}. Also, several recent studies have demonstrated that QML approaches such as quantum support vector machines (QSVMs) and quantum convolutional neural networks can achieve competitive performance with classical machine learning models in classification and prediction tasks~\cite{sha2024, zhuang2024, cheny2024, abdulsalam2025, yang2025}. These studies demonstrate the growing versatility of QML methods across both classification and regression problems, yet a unified comparison framework for QNNs remains underexplored.

As the success of QML's practical implementation in various fields has been proven by recent studies, researchers have worked on the possibility of using QML for classification and prediction purposes in energy studies. While some of the studies have focused on hybrid approaches whose success has already been proven, others have focused on the possibility of fully-quantum approaches. Stability assessment of power systems and smart grids, fault diagnosis of power systems, prediction of power output of renewable energy systems, and predictive management for grid stability were several recent use cases that benefited from the use of QML. In these use cases, QML has served several purposes, such as providing better feature extraction mechanisms and improving the learning capabilities of classical ML systems~\cite{ajagekar2021, zhou2022noise, yu2023prediction, satpathy2024, chen2024, safari2024neuroquman, hangun2024SmartGrid, hangun2024wind}. Although QML-based approaches have shown promise in addressing practical problems within power systems and energy-related applications, there remains a lack of generalized QML models tailored to specific tasks in these domains. Therefore, experimental studies that guide researchers in selecting the most appropriate QML approach for their specific problems are of great importance.

In the current renewable energy scene, wind power stands as a strong alternative to fossil fuels. In 2024, total global installed wind capacity has reached 1 terawatt (TW). By the end of 2029, forecasted capacity will be 2 TW~\cite{gwec2024}. With the growing reliance on wind power, ensuring successful grid integration and maintaining operational efficiency have become critical during wind farm operation. Imbalances between energy supply and demand can pose challenges to integration, while suboptimal power generation scheduling and inefficient utilization of ancillary services may lead to operational inefficiencies. Consequently, accurately forecasting wind farm power output is essential for supporting reliable and efficient energy system operation~\cite{gregor2011, foley2012}. Currently, machine learning represents the primary approach for forecasting wind farm power output, and recent studies highlight the advantages of leveraging artificial intelligence in this context~\cite{ju2019, shabbir2019, sulaiman2024, olcay2024}. 

Although classical ML is a proven approach, the increasing size and complexity (e.g., high dimensionality, complex feature relationships) of data negatively affect the computational and learning performance of ML systems. Built on the principles of quantum mechanics, quantum computing offers unique advantages unattainable by classical computers. Unique quantum concepts, such as entanglement and superposition, enable natural parallelization and enhanced data representation. Hence, researchers aimed to use these advantages brought by quantum computing to solve different problems related to energy forecasting by deploying QML-based approaches. Results from recent studies have shown significant advantages of QML over classical ML for energy-related applications. QML has proved to be a useful approach for problems with complex relationships between the features of the dataset, and problems with high dimensionality~\cite{ajagekar2019, ajagekar2021, zhou2022, zhou2023, ranga2024, hong2025}. 

Recent studies have proved QML's success in the energy domain, but there is still a lack of generalized approaches and/or clear guidance for researchers on selecting the proper quantum approach. However, a knowledge gap remains regarding the optimal quantum circuit structure for specific problems, necessitating further experimental research. Especially, there is limited empirical work evaluating different QNN configurations across dataset scales, making it difficult to identify optimal circuit structures for practical deployment. Such empirical knowledge is critical when it comes to deploying QNNs on real quantum hardware, which is scarce and expensive. Due to limited access to quantum hardware, researchers generally use quantum circuit simulations to test their proposed approaches;   therefore, an evaluation of simulation runtimes on classical hardware is important~\cite{gujju2024}. Studies have reported long simulation running times; therefore, an in-depth investigation of simulation time performance is necessary to provide detailed insights. As a result, accelerating existing quantum approaches and/or benchmarking existing methods to pave the way for efficient implementations is an active focus at the quantum computing field~\cite{budinski2023, smith2023, mineh2023, agliardi2025, ponce2025}. Even though there are studies to show QNN's advantages on energy studies, the field still lacks research that focuses on investigating the different QNN structures for their prediction and simulation time performance. Our previous study has shown that a QNN with a specific data encoding and entanglement strategy can outperform not only other QNN configurations but also compete with classical ML approaches such as k-Nearest Neighbor (kNN), decision tree (DT), and linear regression (LR)~\cite{hangunWind2025}. Based on that work, this study expands it by looking at the problem from a broader perspective that involves data size, time evaluation, and quantum circuit complexity. 

In this study, we present a systematic benchmark of QNN architectures using a regression dataset related to wind turbine power. Specifically, we investigate six different QNN configurations based on variations in ansatz design while employing a Z-feature map. These models are compared against classical regressors such as k-nearest neighbors, decision trees, and linear regression. We also evaluate the ML performance over varying dataset sizes to understand how QNN scalability compares to classical alternatives. Furthermore, we provide a detailed analysis of the asymptotic time complexity for each of the six QNN configurations, examining how their runtime varies with increasing data size. Our aim is to provide practical insights for researchers considering QNN approaches for real-world regression tasks related to wind power studies. 

The rest of this paper is organized as follows: Section~\ref{dataset} describes the dataset and data preparation, Section~\ref{models_and_method} presents experimental design, Section~\ref{exp_res_and_anal} discusses results, and Section~\ref{conclusion} concludes the study and discusses future work.

\section{Dataset and Preprocessing}\label{dataset}
To demonstrate a practical application of quantum computing, this study uses a real-world temporal dataset of a wind turbine's power output. This dataset, previously utilized in our successful applications of classical machine learning regression algorithms for wind power prediction~\cite{eyecioglu2019}, comprises 4464 measurements recorded at 10-minute intervals. Each data point includes four input features that influence power output. The measured parameters include wind speed ($m/s$), wind direction ($\theta$), atmospheric pressure ($hPa$), ambient temperature ($\degree C$), and generated power output ($kW$) of a wind turbine~\cite{DTU_Risoe_WindDatabase}. Basic statistical properties of the dataset are given in Table~\ref{basic_descriptive_stats}. Target feature "\textit{Power}" ranges from 2.24 to 2033.12 kW, which has the widest range of the features.

\begin{table}
\caption{Basic descriptive statistics for the wind turbine dataset.}
\begin{indented}
\item[]\begin{tabular}{@{}lrrrrrr}
\br
Variable    & Mean    & Median  & Std.\ Dev.\ & Min    & Max     & Range\\
\mr
Temperature ($\degree C$) &   3.94  &   4.205 &    2.041    &  -5.29 &   10.00 &   15.29\\
Pressure ($hPa$)   & 1019.46 & 1024.745&   13.054    & 979.79 & 1035.72 &   55.93\\
Direction ($\theta$)   &  243.11 &  247.90 &   55.109    & 100.67 & 359.78  &  259.11 \\
Velocity ($m/s$)     &    8.65 &    7.635&    4.241    &   0.32 &  21.07  &   20.75 \\
Power ($kW$)       &  666.60 &  376.675&  716.757    &   2.24 & 2033.12 & 2030.88 \\
\br
\end{tabular}
\end{indented}
\label{basic_descriptive_stats}
\end{table}

As this study aims to conduct an experimental analysis of dataset size and ansatz configurations, we created four subsets from the original dataset. To better observe how results vary with increasing dataset size, instead of using all 4464 samples, we randomly generated subsets containing 1000, 2000, 3000, and 4000 samples, and used these subsets for each experiment. For each experiment, we divided the data into training (80\%) and test (20\%) sets, and then used a min-max scaler (Equation~\eref{eq:min_max}) to scale all values to the range $[0, 1]$.

\begin{eqnarray}
    x_{scaled}=\frac{x_{original}-x_{min}}{x_{max}-x_{min}}
\label{eq:min_max}
\end{eqnarray}

\section{Methodology}\label{models_and_method}
\subsection{Quantum Machine Learning}\label{qml}
\begin{figure}
    \centering
    \includegraphics[width=0.8\textwidth]{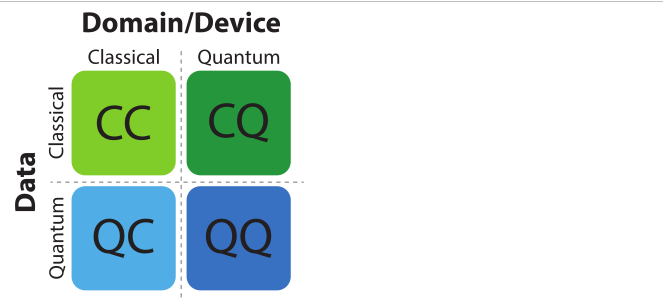}
    \caption{Classification of QML approaches based on data type and processing environment.}
    \label{fig:qml_data_alg}
\end{figure}

QML offers advantages over classical ML by combining principles of quantum computing with classical ML approaches. Such advantages include the ability to handle complex datasets and achieve faster data processing with natural parallelization~\cite{Schuld2015}. Current QML research focuses on combining quantum computing with classical approaches. This focus aims to benefit from the advantages of classical ML and quantum computing at the same time. Depending on the type of data processed and the environment where the data was processed, there are four options~\cite{ranga2024}, as shown in~\ref{fig:qml_data_alg} which are:

\begin{itemize}
    \item Classical-Classical (CC): Classical datasets used with classical approaches. 
    \item Classical-Quantum (CQ): Classical datasets used with quantum approaches.
    \item Quantum-Classical (QC): Quantum datasets used with classical approaches.
    \item Quantum-Quantum (QQ): Quantum datasets used with quantum approaches.
\end{itemize}

Even in the current NISQ era of quantum computing, quantum advantage is still clear to be seen for certain tasks~\cite{preskill2015}. As a result, studies relying on the applications of QML for real-world cases yield great importance to provide insights that will pave the way for the standards of the useful QML. In this study, we focused on using QNN, a specific type of QML, for a prediction. task related to a renewable energy scenario.

\subsection{Quantum Neural Networks}\label{qnn}
\begin{figure}
    \centering
    \includegraphics[width=0.9\linewidth]{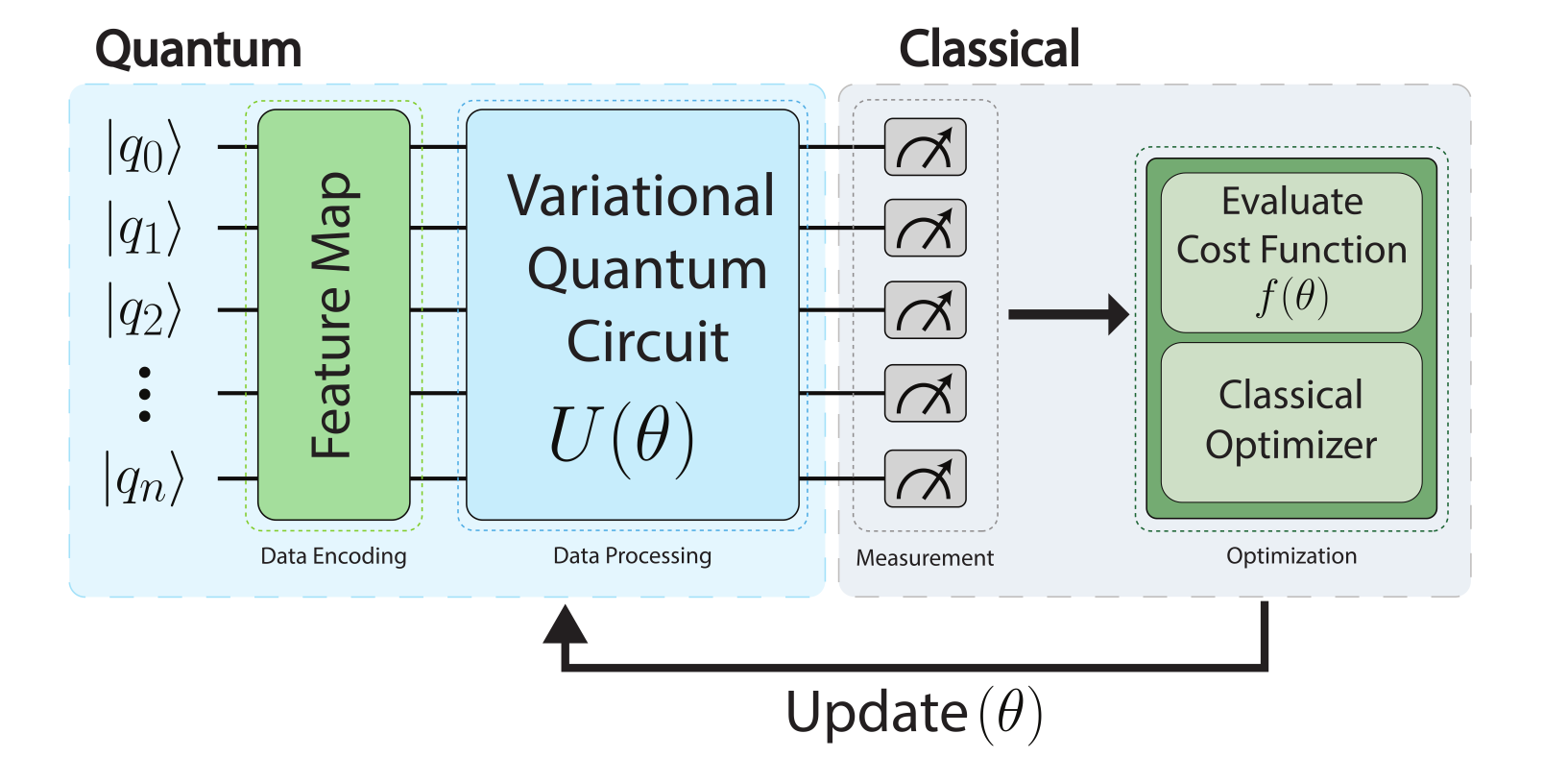}
    \caption{General architecture of the Quantum Neural Network (QNN) used in this study.}
    \label{fig:general_qnn}
\end{figure}

QNN is a hybrid quantum-classical ML approach that suggests a quantum mechanical version of the standard neural networks. It is a variational quantum circuit (VQC) with tunable parameters. A conventional QNN whose general structure was given in Figure\ref{fig:general_qnn} consists of a quantum layer and a classical layer. While the quantum layer works as a part to convert classical data into quantum data and process the quantum data to extract classification/prediction results, the classical layer works as an optimizer to tune the quantum layer so that it can find the best parameters that can minimize or maximize the cost function. 

Even though the internal design of the quantum layer and classical layer depends on the requirements of the specific real-world application, every QNN consists of the generic parts that serve various needs, including:

\begin{enumerate}
    \item Data encoding
    \item Data processing
    \item Measurement
    \item Optimization
\end{enumerate}

\subsubsection{Data Encoding}\label{data_encoding}
Data encoding corresponds to converting the data represented in the classical domain as integers and/or floating-point numbers, or strings to quantum domain representation~\cite{rath2024quantum}. In the QML literature, quantum circuits that take classical data as input to create a quantum representation of it are called \textit{feature maps}. Feature maps encode the classical data using three fundamental approaches:

\begin{enumerate}
    \item Angle Encoding
    \item Amplitude Encoding 
    \item Basis Encoding
\end{enumerate}

In this study, we used a special quantum circuit that is used for data encoding called the \textit{Z Feature Map}. Specifics of that feature map will be provided in Section~\ref{z_feature_map}.

\subsubsection{Data Processing}\label{data_processing}
After classical data is encoded into quantum states, it is ready for processing by quantum circuits. The data processing step uses special quantum circuits called parameterized quantum circuits (PQC), also known as a VQC. This circuit has layers of quantum gates that have tunable parameters (analogous to weights in classical neural networks) that are optimized during the training step. Generally, the structure of the VQC is called \textit{ansatz}. From a quantum mechanical perspective, an ansatz is used as a trial wave function or state that acts as an initial approximation to solve a given quantum problem. Parameters of the ansatz represent the quantum state that is being approximated. With an iterative execution, it is aimed to find optimal parameters that can closely approximate the target quantum state~\cite{yuxuan2022}. As a result, designing a proper ansatz can significantly affect the performance of the QNN. According to research, the choice of circuit architecture can change the model's ability to learn and generalize in both positive and negative ways. In addition to that, due to the NISQ era, researchers need to test newly developed quantum approaches on simulators first. The total time required to simulate a QNN is also affected by the ansatz configuration. This study aims to survey the real-world usability of QNNs under the ansatz perspective. Detailed explanations about ansatz will be given at the Section~\ref{ansatz_configs}.

\subsubsection{Measurement}\label{measurement}
Upon completion of the data processing step, the results must be measured to adjust the VQC parameters.  It is necessary to obtain the results as classical information so that they can be used in the optimization step to update the cost function. The measurement step serves this cause to get results from the quantum domain and then convert them to classical information. 

\subsubsection{Optimization}\label{optimization}
The final step of training a QNN is to use an optimization technique to update the parameters of the VQC that can minimize the difference between the actual values \(y\) and the predicted values \(\hat{y}\). Generally, the optimization step is executed by classical hardware elements (CPUs, GPUs, etc.), classical optimization techniques such as adaptive moment estimation (ADAM) and limited-memory Broyden–Fletcher–Goldfarb–Shanno bound optimizer (L-BFGS-B) are used used for finding optimal parameters for the VQCs at the data processing step. At each iteration, according to results from the optimizer, \(\theta\) the parameter of the VQC is tuned.

\subsection{QNN Architectures}\label{qnn_archs}
In this study, we evaluated six distinct Quantum Neural Network (QNN) configurations based on a specific quantum feature map and varying ansatz strategies. Our previous experimental study~\cite{hangunWind2025}, found that the Z Feature Map demonstrated approximately three times better predictive performance than the ZZ Feature Map on the same dataset used here. Consequently, we chose to concentrate on the Z Feature Map for more comprehensive experimentation. Each configuration involved the Z Feature Map combined with different entanglement strategies for the ansatz. A summary of these QNN configurations is provided in Table~\ref{qnn_configs}. Additionally, to facilitate the reproducibility of our experiments, a visual depiction of the quantum circuits implementing the QNN configurations is included in~\ref{append1}, Figures~\ref{fig:qnn1_circ},~\ref{fig:qnn2_circ},~\ref{fig:qnn3_circ},~\ref{fig:qnn4_circ},~\ref{fig:qnn5_circ},~\ref{fig:qnn6_circ}.

\begin{table}[h]
\centering
\caption{QNN configurations used in this experiments}\label{qnn_configs}%
\begin{tabular}{@{}lll@{}}
    QNN Configuration & Feature Map  & Ansatz \\
    \mr
    QNN-1    & Z Feature Map   & Full   \\
    QNN-2    & Z Feature Map   & Linear    \\
    QNN-3    & Z Feature Map   & Circular    \\
    QNN-4    & Z Feature Map   & SCA    \\
    QNN-5    & Z Feature Map   & Reverse Linear    \\
    QNN-6    & Z Feature Map   & Pairwise    \\
    \br
\end{tabular}
\end{table}

\subsubsection{Z Feature Map}\label{z_feature_map}
In the QML, feature maps are essential to encode classical data into its high-dimensional Hilbert space representation. This representation allows leveraging quantum mechanical phenomena to benefit from the unique advantages of quantum computing, which can't be provided by classical computing. Feature maps in quantum computing enable the implementation of QML approaches like kernel-based methods or QNNs that are proven to be a strong alternative for classical ML approaches. Similar to feature maps in classical ML that can embed the data into high-dimensional spaces, quantum feature maps convert classical data \textit{x} to a quantum state \(\ket{\phi(x)}\) which relies on Hilbert space~\cite{schuld2019}. This type of embedding, also known as encoding, is an essential part of QNNs since the Z feature map serves as a fixed data encoding layer. 

Z feature maps are built using \textit{Hadamard} gates, and \textit{Z-axis rotation} (also known as \textit{phase shift}) gates. Matrix representation of Hadamard (\(H\)) gate, and Z-axis rotation (\(R_Z\)) are given as follows:

\begin{eqnarray}
    H = \frac{1}{\sqrt{2}}
    \left[
      \begin{array}{cc}
        1 & 1 \\
        1 & -1
      \end{array}
    \right]
\end{eqnarray}

\begin{eqnarray}
    R_z(\theta)  = 
    \left[
        \begin{array}{cc}
        e^{\frac{-i\theta}{2}}  & \hspace{1.5em} 0 \\
        0 \hspace{1.5em} &  e^{\frac{i\theta}{2}}  \\
        \end{array}
\right]
\end{eqnarray}

where \(\theta\) is the rotation angle along the Z-axis. Interchangeably, phase shift gate (\(P\)) can also be used instead of the \(R_Z\) gate to obtain the same results. Especially, in Qiskit~\cite{qiskit2024}, \(P\)-gate is the preferred gate for default Z feature map implementation. The matrix representation of the \(P\)-gate is given as follows:

\begin{eqnarray}
    P  = 
    \left[
        \begin{array}{cc}
        1 &  0 \\
        0 &  e^{i\varphi}  \\
        \end{array}
\right]
\end{eqnarray}

where \(\varphi\) represents a rotation angle along the Z-axis on the Bloch sphere.

For a given classical data vector (in ML sense a given feature vector) \(x=(x_1, \; x_2, \; \dots, \; x_n)\), the Z feature map acts on \textit{n} qubits as follows to create quantum states:

\begin{enumerate}
    \item Apply Hadamard gates: $H^{\otimes n}$ to create a superposition.
    \item Apply single-qubit $Z$-axis rotations: $R_Z(2x_j)$ to each qubit $j$.
\end{enumerate}

After applying the given steps, we obtain in the quantum state \(\ket{\psi(x)}\) as follows:

\begin{eqnarray}
    \ket{\psi(x)} = \bigotimes_{j=1}^n \frac{1}{\sqrt{2}} (\ket{0} + e^{i 2x_j} \ket{1})
\end{eqnarray}

We can also represent the Z feature map in terms of quantum gates as follows:

\begin{eqnarray}
    \ket{\psi(x)} = \prod_{j=1}^{n}H_{j}R_z(2x_j)_jH_j 
\end{eqnarray}

To facilitate the reproducibility of our experiments, a visual depiction of the quantum circuit implementing the Z Feature Map is provided in~\ref{append1}. When constructing the Z Feature Map—or quantum circuits in general—it is common to iteratively append identical circuit layers, referred to as repetitions, to achieve more complex interference patterns~\cite{schuld2019}. In this study, we utilized two repetitions, as illustrated in Figure~\ref{fig:appendix_z_feature_map_circ}. Notably, the Z Feature Map does not incorporate entangling gates; thus, the resulting quantum circuit remains shallow and highly hardware-efficient. This characteristic makes it particularly suitable for execution on quantum hardware as well as for efficient classical simulations of quantum circuits. Owing to this simplicity, the Z Feature Map is frequently employed in scenarios where low-depth circuits are essential~\cite{schuld2019,havlicek2019}.

\subsubsection{Ansatz Configurations}\label{ansatz_configs}
In quantum computing, especially for variational quantum algorithms (VQAs), the term \textit{ansatz} refers to a quantum circuit that has tunable parameters that aim to prepare a trial wavefunction \(\ket{\psi(\theta)}\). At that point, parameter \(\theta\) denotes a tunable set of parameters. In problems like finding the ground state energy of a quantum system, tunable parameters are optimized to approximate the solution~\cite{mcclean2016}.

Mathematically, the ansatz is a unitary operator $U(\boldsymbol{\theta})$ acting on an initial reference state represented as $\ket{\psi_0}$, typically chosen as the all-zero state. Using the reference state and unitary operator, tunable parameters of the trial wavefunction are updated using Equation~\eref{eq:psi_theta_ket}.

\begin{eqnarray}
    \ket{\psi(\theta)} = U(\theta)\ket{\psi_0} 
\label{eq:psi_theta_ket}
\end{eqnarray}

The ansatz $U(\boldsymbol{\theta})$ is constructed by connecting parameterized quantum gates such as rotational gates and entangling gates in sequential layers. A generalized form to create an ansatz using alternating layers of two-qubit entangling gates and single-qubit rotational gates was given at Equation~\eref{eq:ansatz_circ_eq}.

\begin{eqnarray}
    U(\theta) = \prod_{l=1}^{L}\Biggl( \bigotimes_{j=1}^{n}R_{\phi j}^{(j)}(\theta_{l,j)}E_{l} \Biggr)
\label{eq:ansatz_circ_eq}
\end{eqnarray}

where $R_{\phi_j}^{(j)}(\theta_{l,j})$ denotes a rotation around the axis $\phi_j \in {x, y, z}$ on qubit $j$ with an angle $\theta_{l,j}$, and $E_l$ represents the entangling operation in layer $l$. 

In our experimental study, we used \textit{Real Amplitude} ansatz, which is widely used in QNN architectures~\cite{abbas2021, yuxuan2022, arthur2022_hybrid}. In the Real Amplitude ansatz, the \textit{Y-axis rotational} gate ((\(R_y\))) is used for rotation operations, and the \textit{Controlled NOT} ((\(CNOT\))) gate is used for the entangling operation. \(R_y\) gate is represented in matrix form given in Equation~\eref{eq:ry_gate}, and \(CNOT\) gate that acts on two qubits is represented in matrix form given in Equation~\eref{eq:cnot_gate}.

\begin{eqnarray}
     R_y(\theta)  = 
    \left[
        \begin{array}{cc} 
        \cos(\frac{\theta}{2}) & -\sin(\frac{\theta}{2}) \\
        \sin(\frac{\theta}{2}) & \cos(\frac{\theta}{2}) \\
        \end{array}
    \right]
\label{eq:ry_gate}
\end{eqnarray}

\begin{eqnarray}
    CNOT  = 
    \left[
        \begin{array}{cccc}
        1 & 0 & 0 &  0\\
        0 & 1 & 0 &  0\\
        0 & 0 & 0 &  1\\
        0 & 0 & 1 &  0\\
        \end{array}
\right]
\label{eq:cnot_gate}
\end{eqnarray}

For the ansatz, the goal is to find the set of parameters $\boldsymbol{\theta}$ that minimizes the expectation value of a cost Hamiltonian $H$ given as Equation~\eref{eq:cost_hamil}.

\begin{eqnarray}
    E(\theta) = \bra{\psi(\theta)}H\ket{\psi(\theta)}
\label{eq:cost_hamil}
\end{eqnarray}

The structure of an ansatz is critical since it affects the expressibility of the quantum circuit, and the efficiency of the algorithms~\cite{mcclean2016}. In this experimental study, we built six different ansätze by changing entanglement strategies. These strategies are listed as, \textit{Full}, \textit{Linear}, \textit{Circular}, \textit{Shifted-circular-alternating (SCA)}, \textit{Reverse linear}, \textit{Pairwise}.

Entanglement strategies define the number and direction of the \(CNOT\) gates that are used to create entanglement between two qubits. The effects of the entanglement strategy on time complexity and on the ML performance of the quantum circuit are factors that need to be investigated experimentally to provide insights to researchers that aim to apply QML in real-world scenarios. We aimed to provide such insight by investigating these effects from different perspectives. Quantum circuit representations of different ansätze used in this study are given in~\ref{fig:full_ansatz_circ},~\ref{fig:linear_ansatz_circ},~\ref{fig:circular_ansatz_circ},~\ref{fig:sca_ansatz_circ},~\ref{fig:reverse_linear_ansatz_circ}. 

Types of the quantum gates depending on the number of qubits they manipulate affect the implementability of quantum circuits on quantum hardware, and the time complexity of the simulation of quantum circuits on classical hardware. In the Table~\ref{qnn_gate_counts}, we provided the total number of single-qubit gates, two-qubit gates, and the total number of gates for each configuration. QNN-1 has the highest number of total gates with 40 gates; on the other hand, QNN-2, QNN-5, and QNN-6 have the least number of total gates with 34 gates.

\begin{table}[h]
\caption{Quantum gate counts for each QNN configuration.}
\label{qnn_gate_counts}
\begin{indented}
\item[]
\footnotesize
\begin{tabular}{@{}lllccr}
\br
QNN Config & Feature Map & Ansatz &
\begin{tabular}[c]{@{}c@{}}Single-Qubit Gates\\(Feature + Ansatz)\end{tabular} &
\begin{tabular}[c]{@{}c@{}}Two-Qubit Gates\\(Feature + Ansatz)\end{tabular} &
\begin{tabular}[c]{@{}c@{}}Total\\Gates\end{tabular} \\
\mr
QNN-1 & Z & Full            & 16 + 12 = 28 & 0 + 12 = 12 & 40 \\
QNN-2 & Z & Linear          & 16 + 12 = 28 & 0 + 6 = 6   & 34 \\
QNN-3 & Z & Circular        & 16 + 12 = 28 & 0 + 8 = 8   & 36 \\
QNN-4 & Z & SCA             & 16 + 12 = 28 & 0 + 8 = 8   & 36 \\
QNN-5 & Z & Reverse Linear  & 16 + 12 = 28 & 0 + 6 = 6   & 34 \\
QNN-6 & Z & Pairwise        & 16 + 12 = 28 & 0 + 6 = 6   & 34 \\
\br
\end{tabular}
\end{indented}
\end{table}

\subsection{Classical Models}\label{classical_models}
For the classical regression benchmarks, we selected the k-Nearest Neighbors (kNN), Decision Tree Regressor (DTR), and Linear Regression (LR) algorithms. This selection was based on our prior experimental success with these specific models on the same wind turbine power output dataset. While detailed implementation specifics for these classical algorithms are available in our previous work, and interested researchers are directed to consult~\cite{eyecioglu2019} for further information, we outline the specific parameter configurations used in the current study. Linear Regression (LR) and Decision Tree Regressor (DTR) algorithms were applied with their default parameters. For the kNN algorithm, the number of neighbors ($k$) was set to 5, and the distance metric employed was Minkowski. Classical models were also experimentally tested using the same training and test dataset regarding the increasing dataset size.

\section{Experimental Results and Discussion}\label{exp_res_and_anal}
This section presents and analyzes the empirical results from our experiments. Initially, we examine six distinct Quantum Neural Network (QNN) configurations, providing a detailed assessment of their machine learning performance and the time efficiency of their simulations. Subsequently, we evaluate three classical Machine Learning (ML) models—k-Nearest Neighbors (kNN), Decision Tree Regressor (DTR), and Linear Regression (LR)—whose efficacy was demonstrated in our prior work~\cite{eyecioglu2019}. A core component of this section is a comparative analysis of the QNNs against these classical ML benchmarks, with particular attention to scalability and generalization capabilities across varying dataset sizes.

The primary metrics for comparing model performance are the coefficient of determination ($R^2$) and Root Mean Square Error $RMSE$. To further elucidate the predictive capabilities of each model, we provide visual comparisons of actual versus predicted values. Finally, error distribution histograms are presented to offer insights into the magnitude and distribution of prediction errors for each model.

The primary aim of these experiments is to demonstrate that QNN configurations utilizing a Z feature map and diverse ansatz structures are a viable alternative to established classical ML approaches for prediction tasks on real-world datasets. The study utilizes a wind turbine power generation dataset, which includes four input features (temperature, pressure, wind direction, and velocity) and one target variable (generated wind power). All experiments were conducted on HP Z1 All-in-One G2 Workstation with 16 GB RAM, Intel\textsuperscript{\textregistered} Xeon\textsuperscript{\textregistered} Processor E3-1246 v3, utilizing Qiskit Machine Learning~\cite{qiskitml2025} within a Python programming environment.

\subsection{Quantum Neural Network Machine Learning Performance}
Our experimental work began with a detailed investigation of machine learning predictive performance. Importantly, this evaluation was consistently carried out using varying dataset sizes to highlight the influence of dataset size on this key performance metric. To evaluate the machine learning performance precisely, we used $R^2$ and $RMSE$ as evaluation metrics. This detailed analysis is followed by an investigation into the learning convergence of QNN configurations using their loss functions. For all QNN configurations, we used L-BFGS-B with 25 iterations, and $\epsilon$(error) of $1e-8$ which is the default value. 

\subsubsection{QNN Cross-Validation: Performance and Stability}
For a comprehensive evaluation of QNN configurations, we utilized training datasets with 800, 1600, 2400, and 3200 samples; these were derived from our original datasets having total sample sizes of 1000, 2000, 3000, and 4000, respectively. A 5-fold cross-validation scheme was employed on each of these training sets for robust model evaluation. The value of \textit{k} was set to five, primarily due to the computational cost associated with QNN simulations. The performance of each QNN configuration, as determined through this cross-validation process, is detailed in Tables~\ref{r2_800_table},~\ref{rmse_800_table},~\ref{r2_1600_table},~\ref{rmse_1600_table},~\ref{r2_2400_table},~\ref{rmse_2400_table},~\ref{r2_3200_table},~\ref{rmse_3200_table}. These tables present the mean $R^2$ and $RMSE$ values, along with their corresponding standard deviations, calculated from the five validation folds for each respective training set size. The standard deviations, in particular, are reported to highlight the stability and robustness of the QNN configurations.

As shown in Tables~\ref{r2_800_table},~\ref{rmse_800_table}, the highest $R^2$ scores on a per-fold basis (0.95) were achieved by QNN-1, QNN-3, and QNN-4, while QNN-1 also recorded the lowest individual fold $R^2$ score (0.85). On average, QNN-1, QNN-2, and QNN-6 performed best among all configurations with an R$^2$ score of 0.91, whereas QNN-3 and QNN-5 achieved the lowest average $R^2$ score of 0.89. Although QNN-2 and QNN-6 were among the top-performing configurations regarding the $R^2$ score, their $R^2$ standard deviation ($\pm$0.031) was higher than that of QNN-5 ($\pm$0.028). For the $RMSE$ during cross-validation, QNN-1 yielded the lowest average error of 210.24 kW. Conversely, QNN-5 achieved the smallest standard deviation in the $RMSE$ across folds, at $\pm$30.01 kW. These results indicate that each configuration provides consistent performance on this smaller dataset.

\begin{table}[h]
    \caption{5-fold cross-validation $R^2$ scores for each QNN configuration on a training set of 800 samples.} 
    \label{r2_800_table}
    \begin{tabular}{@{}lcccccc@{}}
    \toprule
    & \multicolumn{6}{c}{$R^2$}\\\cmidrule{2-7}
    Fold & QNN-1  & QNN-2  & QNN-3  & QNN-4  & QNN-5  & QNN-6   \\
    \mr
    1        & 0.95             & 0.94             & 0.95              & 0.95             & 0.91              & 0.93  \\
    2        & 0.90             & 0.89             & 0.89              & 0.89             & 0.89              & 0.89  \\
    3        & 0.85             & 0.86             & 0.86              & 0.86             & 0.86              & 0.86  \\
    4        & 0.91             & 0.92             & 0.88              & 0.92             & 0.93              & 0.92  \\
    5        & 0.92             & 0.91             & 0.86              & 0.92             & 0.92              & 0.93  \\
    \textbf{Avg.}  & \textbf{0.91 $\pm$0.037} & \textbf{0.91 $\pm$0.031} & \textbf{0.89 $\pm$0.038}  & \textbf{0.89 $\pm$0.050} &\textbf{ 0.90 $\pm$0.028}   & \textbf{0.91 $\pm$0.031}  \\
    \br
\end{tabular}
\end{table}

\begin{table}[h]
\caption{5-fold cross-validation $RMSE\ (\mathrm{kW})$ scores for each QNN configuration on a training set of 800 samples.}
\label{rmse_800_table}
\footnotesize
\begin{tabular}{@{}lcccccc}
\br
& \multicolumn{6}{c}{$RMSE$} \\
\mr
Fold & QNN-1 & QNN-2 & QNN-3 & QNN-4 & QNN-5 & QNN-6 \\
\mr
1 & 154.41 & 173.51 & 156.59 & 166.57 & 214.29 & 185.65 \\
2 & 235.62 & 244.94 & 243.87 & 315.51 & 241.63 & 244.30 \\
3 & 262.95 & 256.57 & 256.88 & 258.16 & 256.61 & 256.16 \\
4 & 209.90 & 197.95 & 251.21 & 202.76 & 185.71 & 196.56 \\
5 & 188.31 & 203.52 & 252.26 & 185.75 & 195.61 & 178.91 \\
\textbf{Avg} & \textbf{210.24 $\pm$ 41.89} & \textbf{215.30 $\pm$ 34.52} & \textbf{232.16 $\pm$ 42.50} & \textbf{225.75 $\pm$ 60.70} & \textbf{218.77 $\pm$ 30.01} & \textbf{212.32 $\pm$ 35.42} \\
\br
\end{tabular}
\end{table}

In the second experiment, the dataset size was doubled, resulting in 1600 samples for cross-validation. As shown in Tables~\ref{r2_1600_table},~\ref{rmse_1600_table}, the highest $R^2$ scores on a per-fold basis (0.94) were achieved by QNN-1, QNN-4, QNN-5 and QNN-6, while QNN-2 and QNN-3 recorded the lowest individual fold $R^2$ score (0.88). On average, QNN-1 and QNN-5 performed best among all configurations with an $R^2$ score of 0.93, whereas QNN-3 achieved the lowest average $R^2$ score of 0.91. Although QNN-1 and QNN-5 were among the top-performing configurations regarding the $R^2$ score, their respective $R^2$ standard deviations ($\pm$0.008 and $\pm$0.009) were higher than that of QNN-6 ($\pm$0.005). For the $RMSE$ during cross-validation, QNN-5 yielded the lowest average error of 188.18 kW. Conversely, QNN-6 achieved the smallest standard deviation in $RMSE$ across folds, at $\pm$8.14 kW. These results indicate that increasing the dataset size resulted in a drop in $RMSE$ for all configurations, and an increase in the $R^2$ score for several configurations.


\begin{table}[h]
    \caption{5-fold cross-validation $R^2$ scores for each QNN configuration on a training set of 1600 samples.} 
    \label{r2_1600_table}
    \begin{tabular}{@{}lcccccc@{}}
    \toprule
    & \multicolumn{6}{c}{$R^2$}\\\cmidrule{2-7}
    Fold & QNN-1  & QNN-2  & QNN-3  & QNN-4  & QNN-5  & QNN-6   \\
    \mr
    1                & 0.93             & 0.91             & 0.89              & 0.92             & 0.94              & 0.92  \\
    2                & 0.94             & 0.93             & 0.93              & 0.94             & 0.94              & 0.93  \\
    3                & 0.92             & 0.92             & 0.93              & 0.90             & 0.93              & 0.92  \\
    4                & 0.94             & 0.93             & 0.93              & 0.93             & 0.94              & 0.93  \\
    5                & 0.92             & 0.88             & 0.88              & 0.92             & 0.92              & 0.92  \\
    \textbf{Avg} & \textbf{0.93 $\pm$0.008}  & \textbf{0.92 $\pm$0.023}  & \textbf{0.91 $\pm$0.025}   & \textbf{0.92 $\pm$0.014}  & \textbf{0.93 $\pm$0.009}   & \textbf{0.92 $\pm$0.005}  \\
    \br
\end{tabular}
\end{table}

\begin{table}[h]
\caption{5-fold cross-validation $RMSE\ (\mathrm{kW})$ scores for each QNN configuration on a training set of 1600 samples.}
\label{rmse_1600_table}
\footnotesize
\begin{tabular}{@{}lcccccc}
\br
& \multicolumn{6}{c}{$RMSE$} \\
\mr
Fold & QNN-1 & QNN-2 & QNN-3 & QNN-4 & QNN-5 & QNN-6 \\
\mr
1 & 199.35 & 213.96 & 243.42 & 207.34 & 183.85 & 205.50 \\
2 & 176.67 & 191.95 & 201.52 & 177.97 & 176.07 & 198.60 \\
3 & 194.10 & 193.22 & 188.69 & 217.76 & 191.28 & 198.96 \\
4 & 182.23 & 185.15 & 184.22 & 188.90 & 181.42 & 187.75 \\
5 & 204.09 & 257.49 & 257.79 & 209.14 & 208.28 & 209.07 \\
\textbf{Avg} & \textbf{191.29 $\pm$ 11.53} & \textbf{208.35 $\pm$ 29.50} & \textbf{215.13 $\pm$ 33.38} & \textbf{200.22 $\pm$ 16.28} & \textbf{188.18 $\pm$ 12.49} & \textbf{199.98 $\pm$ 8.14} \\
\br
\end{tabular}
\end{table}

The third experiment was conducted using a dataset containing 2400 samples, a size triple that of the dataset used in the first experiment. As shown in Tables~\ref{r2_2400_table},~\ref{rmse_2400_table}, the highest $R^2$ score on a per-fold basis (0.96) was achieved by QNN-3, while QNN-2 recorded the lowest individual fold $R^2$ score (0.87). On average, QNN-1, QNN-3, QNN-4 and QNN-5 performed best among all configurations with an $R^2$ score of 0.92, whereas QNN-2 achieved the lowest average $R^2$ score of 0.90. Although QNN-1, QNN-3, QNN-4 and QNN-5 were among the top-performing configurations regarding the $R^2$ score, their respective $R^2$ standard deviations varied; QNN-5 exhibited the lowest standard deviation among these ($\pm$0.015). For the $RMSE$ during cross-validation, QNN-1, QNN-3 and QNN-5 yielded the lowest average error of approximately 201 kW. Although QNN-6 had the highest $RMSE$, it achieved the smallest standard deviation in $RMSE$ across folds, at $\pm$16.43 kW. According to the results, in the third experiment, all QNN configurations achieved an $R^2$ score of at least 0.90; on the other hand, the $RMSE$ tended to increase as the dataset size grew.

\begin{table}[h]
    \caption{5-fold cross-validation $R^2$ scores for each QNN configuration on a training set of 2400 samples.} 
    \label{r2_2400_table}
    \begin{tabular}{@{}lcccccc@{}}
    \toprule
    & \multicolumn{6}{c}{$R^2$}\\\cmidrule{2-7}
    Fold & QNN-1  & QNN-2  & QNN-3  & QNN-4  & QNN-5  & QNN-6   \\
    \mr
    1                & 0.93             & 0.92             & 0.91              & 0.92             & 0.93              & 0.92  \\
    2                & 0.90             & 0.89             & 0.89              & 0.90             & 0.89              & 0.89  \\
    3                & 0.90             & 0.90             & 0.92              & 0.90             & 0.91              & 0.90  \\
    4                & 0.92             & 0.91             & 0.92              & 0.91             & 0.91              & 0.91  \\
    5                & 0.95             & 0.87             & 0.96              & 0.95             & 0.95              & 0.93  \\
    \textbf{Avg} & \textbf{0.92 $\pm$0.020}  & \textbf{0.90 $\pm$0.019}  & \textbf{0.92 $\pm$0.024}   & \textbf{0.92 $\pm$0.018}  & \textbf{0.92 $\pm$0.022}   & \textbf{0.91 $\pm$0.015}  \\
    \br
\end{tabular}
\end{table}

\begin{table}[h]
\caption{5-fold cross-validation $RMSE\ (\mathrm{kW})$ scores for each QNN configuration on a training set of 2400 samples.}
\label{rmse_2400_table}
\footnotesize
\begin{tabular}{@{}lcccccc}
\br
& \multicolumn{6}{c}{$RMSE$} \\
\mr
Fold & QNN-1 & QNN-2 & QNN-3 & QNN-4 & QNN-5 & QNN-6 \\
\mr
1 & 201.57 & 202.76 & 223.13 & 204.87 & 189.24 & 208.11 \\
2 & 222.75 & 228.03 & 231.86 & 220.04 & 227.78 & 232.94 \\
3 & 223.96 & 227.20 & 204.04 & 222.92 & 218.57 & 224.83 \\
4 & 198.11 & 205.67 & 198.41 & 203.41 & 211.44 & 209.19 \\
5 & 158.81 & 255.86 & 150.38 & 164.97 & 162.30 & 190.54 \\
\textbf{Avg} & \textbf{201.04 $\pm$ 26.40} & \textbf{223.90 $\pm$ 21.38} & \textbf{201.56 $\pm$ 31.69} & \textbf{203.24 $\pm$ 23.11} & \textbf{201.87 $\pm$ 26.30} & \textbf{213.12 $\pm$ 16.43} \\
\br
\end{tabular}
\end{table}

The final experiment to assess the overall performance of the six QNN configurations was conducted using a dataset of 3200 samples, a size four times that of the dataset used in the first experiment. As shown in Tables~\ref{r2_3200_table},~\ref{rmse_3200_table}, the highest $R^2$ score on a per-fold basis (0.94) was achieved by QNN-3, while QNN-2 recorded the lowest individual fold $R^2$ score (0.84). On average, QNN-1 and QNN-4 performed best among all configurations with an $R^2$ score of 0.92, whereas QNN-2 achieved the lowest average $R^2$ score of 0.89. Although QNN-1 and QNN-4 were among the top-performing configurations regarding the $R^2$ score, their respective $R^2$ standard deviations varied. Notably, QNN-4 achieved the lowest recorded $R^2$ standard deviation in this experiment, at $\pm$0.010. For the $RMSE$ during cross-validation, QNN-1 yielded the lowest average error of ~205 kW. QNN-4, which had the second-lowest $RMSE$, exhibited the lowest standard deviation of error, at $\pm$11.00 kW. With the exception of QNN-2, the remaining QNN configurations yielded $R^2$ scores above 0.91; however, a slight increase in $RMSE$ was also observed for these configurations.

\begin{table}[h]
    \caption{5-fold cross-validation $R^2$ scores for each QNN configuration on a training set of 3200 samples.} 
    \label{r2_3200_table}
    \begin{tabular}{@{}lcccccc@{}}
    \toprule
    & \multicolumn{6}{c}{$R^2$}\\\cmidrule{2-7}
    Fold & QNN-1  & QNN-2  & QNN-3  & QNN-4  & QNN-5  & QNN-6   \\
    \mr
    1                & 0.91             & 0.88             & 0.92              & 0.92             & 0.91              & 0.91  \\
    2                & 0.92             & 0.91             & 0.92              & 0.91             & 0.91              & 0.91  \\
    3                & 0.90             & 0.84             & 0.86              & 0.90             & 0.88              & 0.89  \\
    4                & 0.93             & 0.89             & 0.94              & 0.92             & 0.93              & 0.92  \\
    5                & 0.93             & 0.93             & 0.92              & 0.93             & 0.93              & 0.91  \\
    \textbf{Avg} & \textbf{0.92 $\pm$0.012}  & \textbf{0.89 $\pm$0.032}  & \textbf{0.91 $\pm$0.028}   & \textbf{0.92 $\pm$0.010}  & \textbf{0.91 $\pm$0.017}   & \textbf{0.91 $\pm$0.013}  \\
    \br
\end{tabular}
\end{table}

\begin{table}[h]
\caption{5-fold cross-validation $RMSE\ (\mathrm{kW})$ scores for each QNN configuration on a training set of 3200 samples.}
\label{rmse_3200_table}
\footnotesize
\begin{tabular}{@{}lcccccc}
\br
& \multicolumn{6}{c}{$RMSE$} \\
\mr
Fold & QNN-1 & QNN-2 & QNN-3 & QNN-4 & QNN-5 & QNN-6 \\
\mr
1 & 212.95 & 246.96 & 209.09 & 209.83 & 216.62 & 217.49 \\
2 & 205.30 & 210.18 & 196.80 & 216.86 & 208.65 & 214.66 \\
3 & 221.50 & 275.62 & 254.09 & 216.19 & 234.36 & 232.45 \\
4 & 194.93 & 240.18 & 181.78 & 200.80 & 193.73 & 199.82 \\
5 & 192.17 & 194.93 & 196.41 & 190.98 & 192.81 & 212.92 \\
\textbf{Avg} & \textbf{205.37 $\pm$ 12.25} & \textbf{233.57 $\pm$ 31.74} & \textbf{207.64 $\pm$ 27.71} & \textbf{206.93 $\pm$ 11.00} & \textbf{209.23 $\pm$ 17.29} & \textbf{215.47 $\pm$ 11.66} \\
\br
\end{tabular}
\end{table}
\newpage
The cross-validation experiments revealed that QNN configurations exhibited varying sensitivities to changes in dataset size. For instance, the $R^2$ score for QNN-1 generally improved with larger datasets up to a certain size, after which the benefit diminished or plateaued. In contrast, QNN-2's $R^2$ score initially increased with more data but then declined as the dataset size grew further. Similarly, QNN-4's $R^2$ score improved with increasing data before stabilizing at larger dataset sizes. A general trend was also observed for the $RMSE$, which typically decreased as the dataset size grew to a certain level, but then tended to increase with further data additions. Furthermore, the standard deviations of both $R^2$ and $RMSE$ exhibited variability across the different dataset sizes, reflecting changes in the stability and robustness of the QNN configurations. The following section will further contextualize these observations by assessing the overall predictive performance of these QNN configurations.

\subsubsection{Overall Predictive Performance}
\begin{figure}
    \centering
    \includegraphics[width=1.1\linewidth]{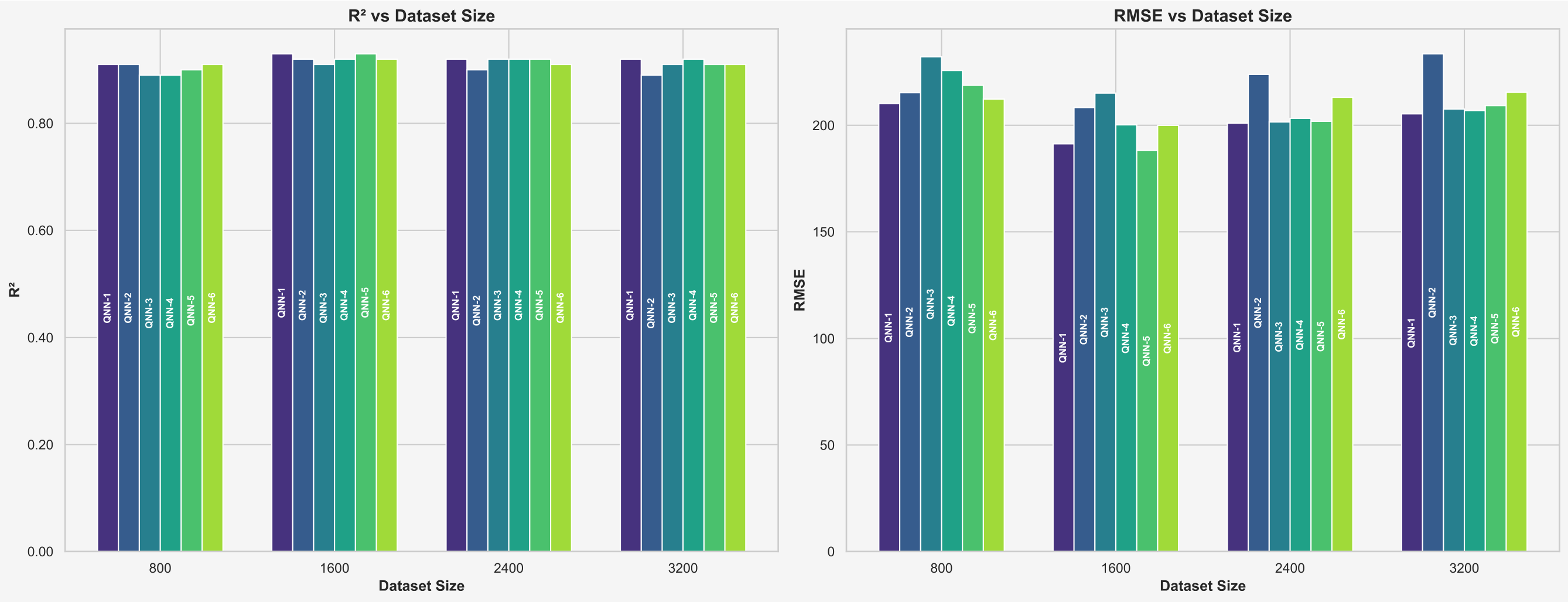}
    \caption{Average $R^2$ and $RMSE$ from cross-validation for each QNN configuration across different dataset sizes.}
    \label{fig:r2_rmse_bar_plot_quantum}
\end{figure}

Figures~\ref{fig:r2_rmse_bar_plot_quantum} depicts the overall predictive performance of the six QNN configurations using $R^2$ and $RMSE$ as performance metrics. Throughout the tests, QNN-1 was consistently among the top-performing configurations in terms of the $R^2$ score. Similarly, for $RMSE$, QNN-1 also demonstrated excellent performance, often yielding the lowest error among the configurations. Regarding the stability of the models, as indicated by the standard deviations of $R^2$ and $RMSE$, no single configuration consistently outperformed others across all dataset sizes, making it difficult to identify a definitive "best" configuration based solely on this aspect.

Notably, improvements in both $R^2$ and $RMSE$ were generally observed as the dataset size increased up to 1600 samples. Beyond this point (1600 samples), further increases in dataset size did not typically yield additional performance gains and, in some cases, led to a decline. For instance, after an initial improvement up to 1600 samples, the $R^2$ scores for QNN-2 and QNN-5 tended to decrease with larger datasets. In contrast, for QNN-1 and QNN-6, $R^2$ scores, after peaking or performing strongly at 1600 samples, slightly decreased at 2400 samples and then largely stabilized for the 3200-sample dataset. Concerning $RMSE$, a general trend of an increase in error was observed for all configurations when the dataset size exceeded 1600 samples. The standard deviations of $R^2$ and $RMSE$ did not exhibit a universally consistent trend across all QNNs. While many configurations showed improved stability (a decrease in standard deviation) as the dataset size approached 1600 samples, their behavior became more varied and less predictable with further increases in data, without a single discernible pattern applicable to all.

It is important to highlight this point of diminishing returns observed at the 1600-sample mark. This transition point occurred when the training data reached 1600 samples, notably half the maximum dataset size (3200 samples) used in these experiments. Future work could involve conducting similar evaluations with a broader range of dataset types and sizes. Such investigations would be valuable to determine if the observed trends, particularly the performance peak at a specific data volume relative to the maximum, represent a generalizable characteristic of these QNN architectures.

\subsubsection{Loss Function Analysis}
\begin{figure}
    \centering
    \includegraphics[width=1.\linewidth]{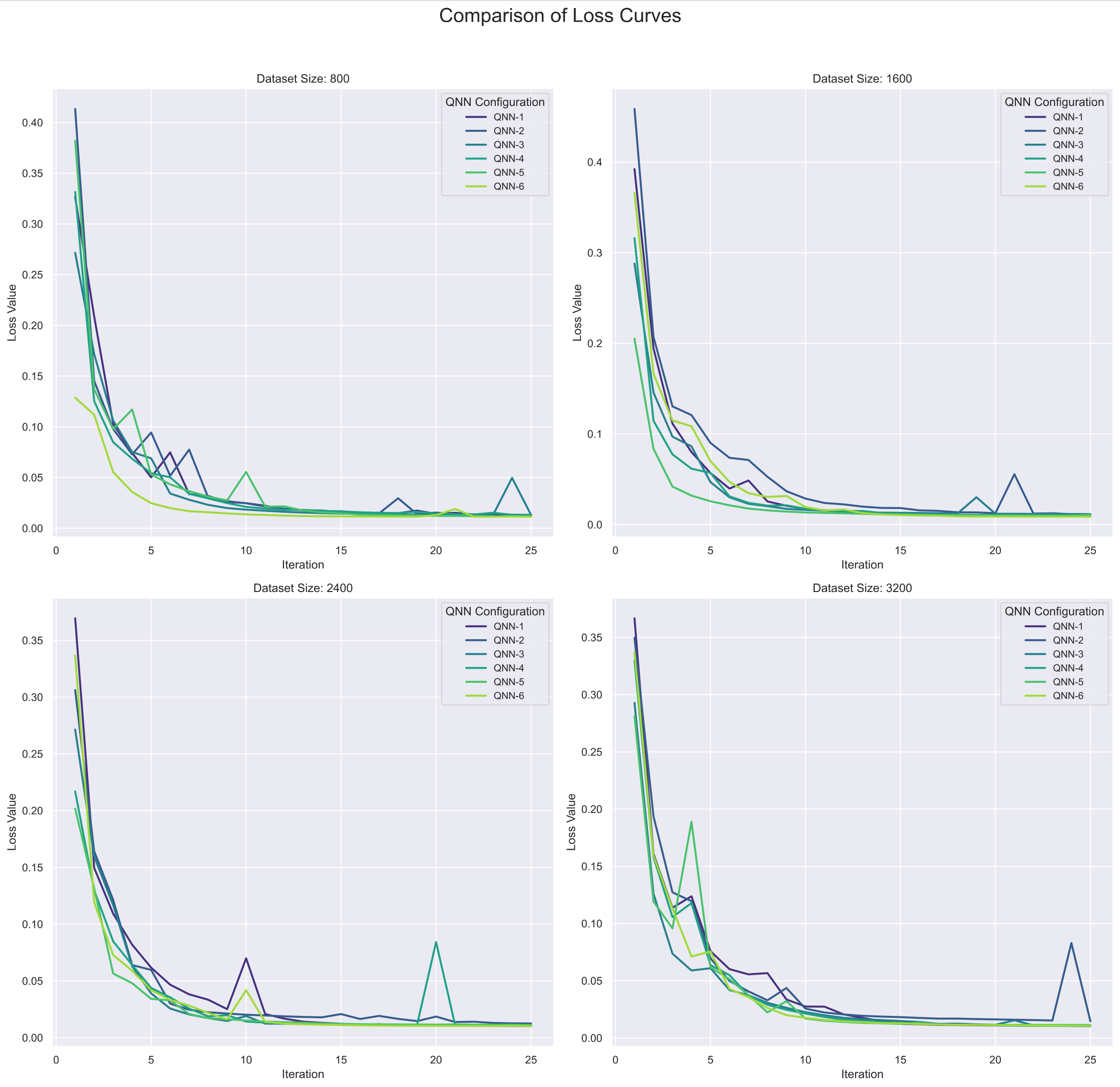}
    \caption{Average training loss per iteration for each QNN configuration across different dataset sizes, based on 5-fold cross-validation.}
    \label{fig:loss_plot}
\end{figure}

In Figure~\ref{fig:loss_plot}, we present training trends for QNN configurations depending on the increasing size of the data set. These loss functions were obtained by averaging the results acquired at each fold of the k-fold cross validation. If we look at the overall training performance, among the all QNN configurations and dataset sizes, the value of the loss function decrease significantly between the approximately first 5-10 iterations. This indicates that all QNN configurations can learn effectively in the early stages of training. After this initial drop, the loss values tend to stabilize, exhibiting only minor fluctuations that emerge as spikes in the plots. This means that for this prediction problem, the QNNs do not require more than approximately 15 iterations to converge. This behavior shows that, dataset size doesn't drastically affect the convergence speed of the learning process. As we mentioned, during the experiments we also noticed some spikes during the training of some QNN configurations. We observed these spikes are not frequent and they do not appear in a regular/noticeable pattern for any of the QNN configurations after the convergence phase ($\approx$10\textsuperscript{th} iteration). For the dataset size 800, we observed a high variance in the loss curves, for example QNN-3 and QNN-5 show sharp spikes. This may indicate overfitting or an unstable training process due to lack of data. For the medium datasets (1600 samples, 2400 samples), even though there are occasional minor spikes, we observe smoother curves, and improved overall convergence with more training samples. For the large dataset (3200 samples), we observed the best stability, and minimal loss function. All of the QNN configurations converged effectively and they maintain low final loss values. It should be noted that, to assess stability, we focused on the change of the loss value after the initial convergence phase which is typically occurs after $\approx$10 iterations. After that point, loss values plateau.

To quantitatively assess the QNN configuration's ability to learn and their stability, we developed an empirical stability score (\textit{SC}) given in Equation~\eref{eq:stability_score}. Stability score was calculated using standard deviation (\textit{SD}) of loss, maximum spike (\textit{MS}) value after 10\textsuperscript{th} iteration, and final loss value (\textit{FL}) at the 25\textsuperscript{th} iteration. We then averaged these values over the four dataset sizes and normalized each metric to the range $[0,1]$ using min-max normalization. To find the stability score (\textit{SC}), we used Equation~\eref{eq:stability_score}. 

\begin{eqnarray}
    \mathrm{SC} = normalized(\mathrm{SD}) + normalized(\mathrm{MS}) + normalized(\mathrm{FL})
    \label{eq:stability_score}
\end{eqnarray}

Table~\ref{stability_score_ranking} summarizes the stability metrics we used to assess the learning performance for each QNN configuration. According to the results, QNN-6 is the most stable model across all metrics and therefore ranks highest overall. It is followed by QNN-5, which exhibits low spikes and low variance. Among all, QNN-1 is a balanced configuration with slightly higher variation. Finally, QNN-2 performs  the worst, with the highest SC.

\begin{table}[h]
\caption{QNN stability ranking based on average standard deviation (SD), maximum spike (MS), and final loss (FL) values. (Lower SC means higher rank)}
\label{stability_score_ranking}
\begin{tabular}{@{}lcccccc@{}}
\toprule
QNN Configuration & SD & MS & FL & SC & Stability Rank \\
\mr
QNN-6 & 0.002 & 0.003 & 0.010 & \textbf{0.07} & 1 \\
QNN-5 & 0.002 & 0.002 & 0.011 & 0.38 & 2 \\
QNN-1 & 0.003 & 0.004 & 0.011 & 0.41 & 3 \\
QNN-3 & 0.004 & 0.015 & 0.011 & 1.26 & 4 \\
QNN-4 & 0.006 & 0.020 & 0.011 & 1.61 & 5 \\
QNN-2 & 0.009 & 0.033 & 0.013 & 3.00 & 6 \\
\br
\end{tabular}
\end{table}

Our detailed experimentation investigated the relationship between dataset size and learning ability/stability of the QNN configurations. Numerical assessment explains that even with a low number of samples, QNNs can learn the data but it comes with the instability of learning. As the dataset size increased, models get higher stability which means when it comes to real-world application, there is a trade-off between model stability and learning ability. It should be noted that, due to the long training times of the QNN configurations, hyper-parameter optimization was not investigated in this study. Hence, for a future work, optimizer related parameter selection might also need to be observed to see if some optimizers are more compatible with specific problems and/or QNN configurations to provide a more stable learning process with low data set size.

\subsection{Quantum Neural Network Simulation Time Performance}
Generally, current implementations of QNNs rely on optimizers that work on classical hardware; as a result, current limitations classical hardware faces are relevant for QNNs. In this section, we investigated the training time for each QNN configuration. In doing so, we examined it from two perspectives: the effect of the dataset size and the effect of the quantum circuit complexity regarding the total number of quantum gates to create each QNN configuration. To evaluate time performance, we used minutes as the unit of time.

\subsubsection{Training Times and Scalability}
Table~\ref{avg_training_times} represents the time evaluation of each QNN configuration over 5-fold cross-validation experiments regarding increasing dataset size. Among all, QNN-1 is the configuration that consistently requires the longest training time independent of the dataset size, on the other hand, QNN-5 and QNN-6 are the most time-efficient configurations among the all of configurations.

\begin{table}[h]
\caption{Average training times (minutes) for QNN configurations over 5-fold cross-validation across different dataset sizes.}
\label{avg_training_times}
\begin{tabular}{@{}lccccc@{}}
\toprule
 & \multicolumn{4}{c}{\textbf{Time (minutes)}} \\
 \cmidrule{2-5}
 & \multicolumn{4}{c}{\textbf{Dataset size}} \\
\cmidrule{2-5}
\textbf{QNN Configuration} & \textbf{800} & \textbf{1600} & \textbf{2400} & \textbf{3200} \\
\mr
QNN-1 & 26.33 & 51.92 & 77.52 & 104.68 \\
QNN-2 & 23.89 & 47.44 & 72.03 & 94.80 \\
QNN-3 & 23.71 & 47.43 & 72.18 & 96.83 \\
QNN-4 & 24.65 & 47.57 & 73.34 & 96.17 \\
QNN-5 & 23.10 & 47.54 & 68.62 & 92.53 \\
QNN-6 & 23.51 & 45.62 & 68.65 & 92.71 \\
\br
\end{tabular}
\end{table}

\begin{figure}
    \centering
    \includegraphics[width=1\linewidth]{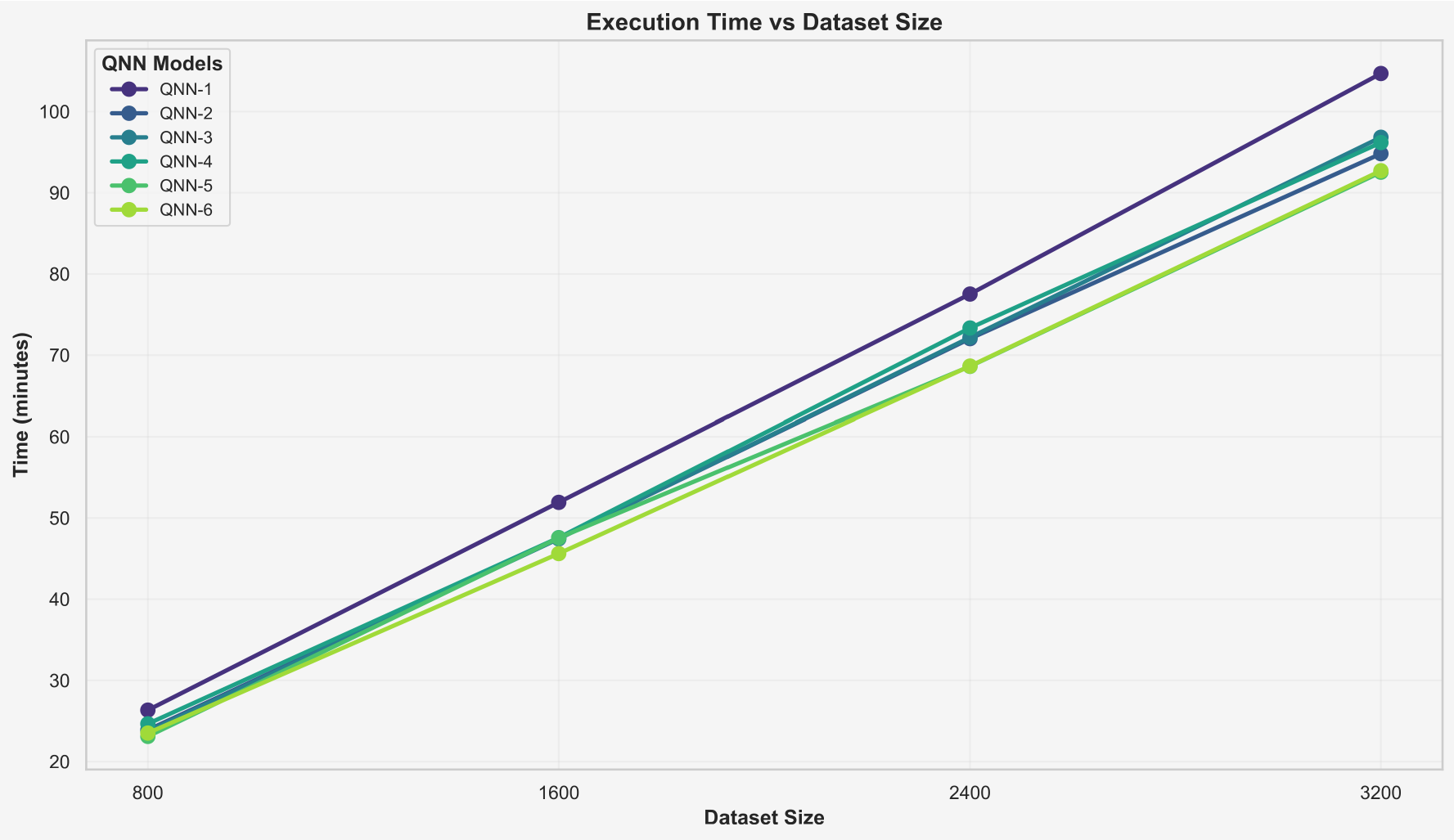}
    \caption{Training time of each QNN configuration as a function of dataset size.}
    \label{fig:time_vs_data_size}
\end{figure}

In addition to Table~\ref{avg_training_times}, we also provided a line plot given in Figure~\ref{fig:time_vs_data_size} to better visualize each QNN configuration's time performance depending on the dataset size. 

To make a better time performance assessment for QNN configurations, we fit a linear regression model given in Equation~\eref{eq:time_perf_model}.

\begin{eqnarray}
    Time = a\times\mathrm{Dataset Size}+b
    \label{eq:time_perf_model}
\end{eqnarray}

Here, the slope $a$ indicates how much time increases per 800 additional samples, which allows us to experimentally quantify the time complexity trend for each QNN configuration and to rank them based on time efficiency. Intercept $b$ is added to the formula to represent initial overheads such as circuit preparation and compilation time. According to this analysis, QNN-5 is the most time-efficient configuration, while QNN-1 is the least efficient.

\begin{table}[h]
\centering
\caption{Time efficiency ranking of QNN configurations based on a linear regression of training time versus dataset size.}
\label{tab:qnn_efficiency_nisq}
\begin{tabular}{lrrr}
\toprule
\textbf{QNN} & \textbf{Slope} & \textbf{Intercept} & \textbf{Rank} \\
\mr
QNN-5 & $2.87e{-02}$ & $0.60$ & 1 \\
QNN-6 & $2.88e{-02}$ & $-0.03$ & 2 \\
QNN-2 & $2.97e{-02}$ & $0.21$ & 3 \\
QNN-4 & $3.00e{-02}$ & $0.35$ & 4 \\
QNN-3 & $3.05e{-02}$ & $-0.99$ & 5 \\
QNN-1 & $3.26e{-02}$ & $-0.05$ & 6 \\
\br
\end{tabular}
\end{table}

Our empirical regression analysis demonstrates that the training of the QNN configurations used in this study scales linearly with dataset size, indicating a time complexity of $\mathcal{O}(n)$.
\subsubsection{Impact of Circuit Complexity}
\begin{figure}
    \centering
    \includegraphics[width=1\linewidth]{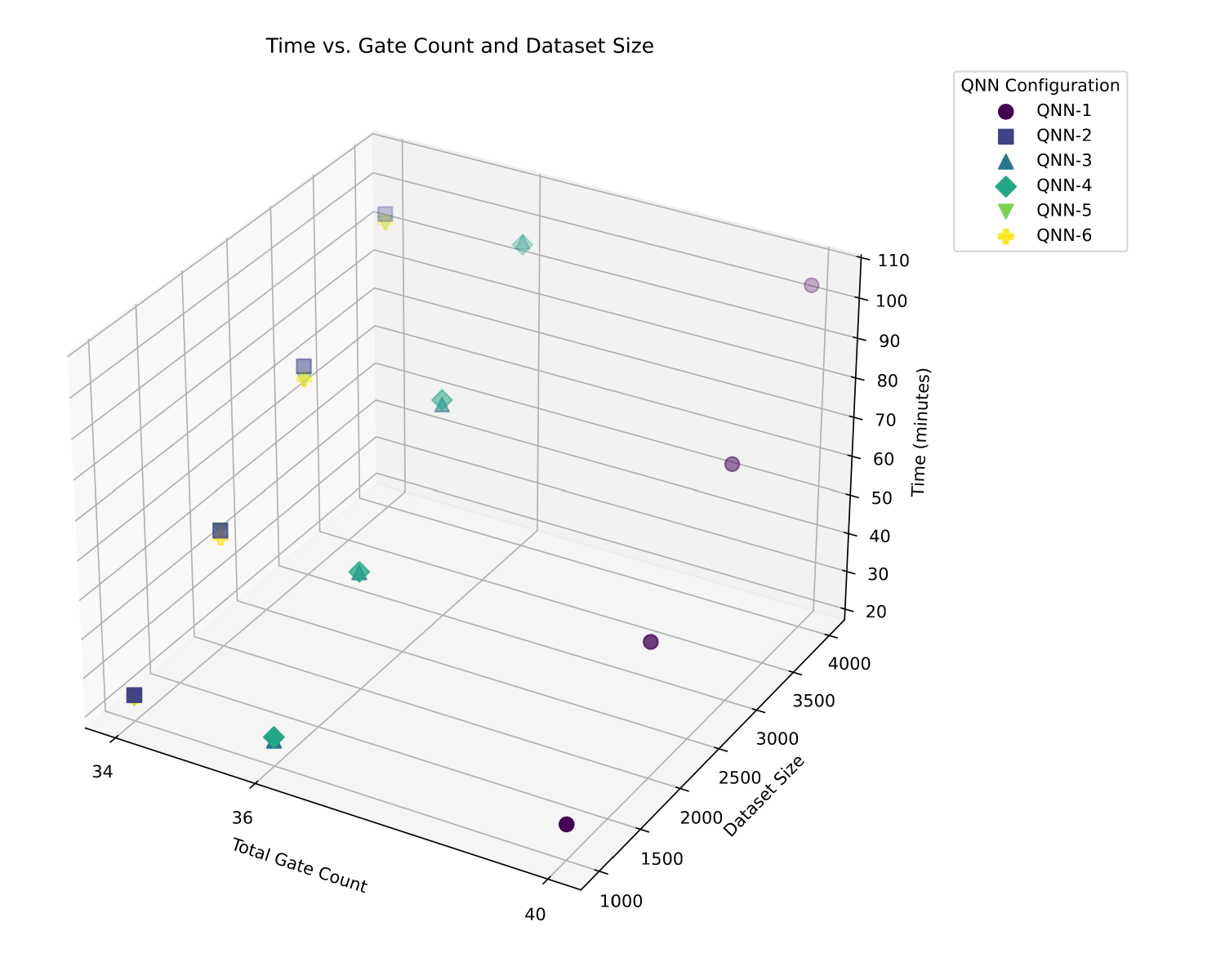}
    \caption{Relationship between total quantum gate count, dataset size, and average training time for each QNN configuration.}
    \label{fig:gc_time_dataset_size}
\end{figure}

In Figure~\ref{fig:gc_time_dataset_size} we provided a compact visualization that shows the effect of total gate count on time performance depending on various dataset sizes. The results show that QNN-1, which has the highest total gate count, exhibits the worst time performance. In contrast, QNN-2, QNN-5, and QNN-6, which have the lowest total gate count, show the best time performance.

The results indicate that QNN configurations with the same total gate count exhibit similar time performance, irrespective of differences in their circuit geometries or qubit entanglement strategies. Furthermore, a linear increase in runtime was observed as the gate count increased across various entanglement strategies. Consequently, the total gate count emerges as a critical parameter that must be considered when designing quantum circuits for QML simulations.

\subsection{Classical Model Performance}
For a fair comparison, we also investigated the classical ML approaches that have proven their success in our previous work~\cite{eyecioglu2019}. Similar to QNN configurations, we tested classical approaches under varying dataset sizes. To evaluate the machine learning performance precisely, we used $R^2$ and $RMSE$ as evaluation metrics. In Tables~\ref{combined_metrics_800_table_classic},~\ref{combined_metrics_1600_table_classic},~\ref{combined_metrics_2400_table_classic},~\ref{combined_metrics_3200_table_classic}, we provided k-fold cross-validation results for classical models k-Nearest Neighbor regression (kNN), decision tree regression (DTR), and linear regression (LR). According to results, for all dataset sizes, DTR and kNN models outperform the LR model in terms of both average $R^2$ and $RMSE$, which indicates stronger predictive capability and lower error margins. 

At the small dataset size (800), LR achieved the best $R^2$ but was followed closely by DTR. Similarly, LR achieved the best $RMSE$ with lowest variance, and followed closely by DTR. For the medium dataset size (1600), DTR takes the lead from LR with an $R^2$ of 0.93 and an $RMSE$ of 193.78 kW. Although DTR achieves the highest $R^2$ and lowest RMSE, kNN exhibits the lowest $RMSE$ variance among the models. In the second middle dataset size (2400), DTR again is the best model in terms of $R^2$ and $RMSE$ of 0.91 and 208.66 respectively. On the other hand, kNN has the lowest variance of $R^2$ and $RMSE$ among all models. Finally, for the large dataset size (3200), kNN achieves both highest $R^2$ of 0.92, and lowest $RMSE$ of 204.18 among all models. It also has the lowest variance of $R^2$ by $\pm0.013$. Even though LR has the lowest $R^2$ of 0.88, and highest $RMSE$ of 247.36, it has the lowest variance of $RMSE$ by $\pm11.00$.

\begin{figure}
    \centering
    \includegraphics[width=1.1\linewidth]{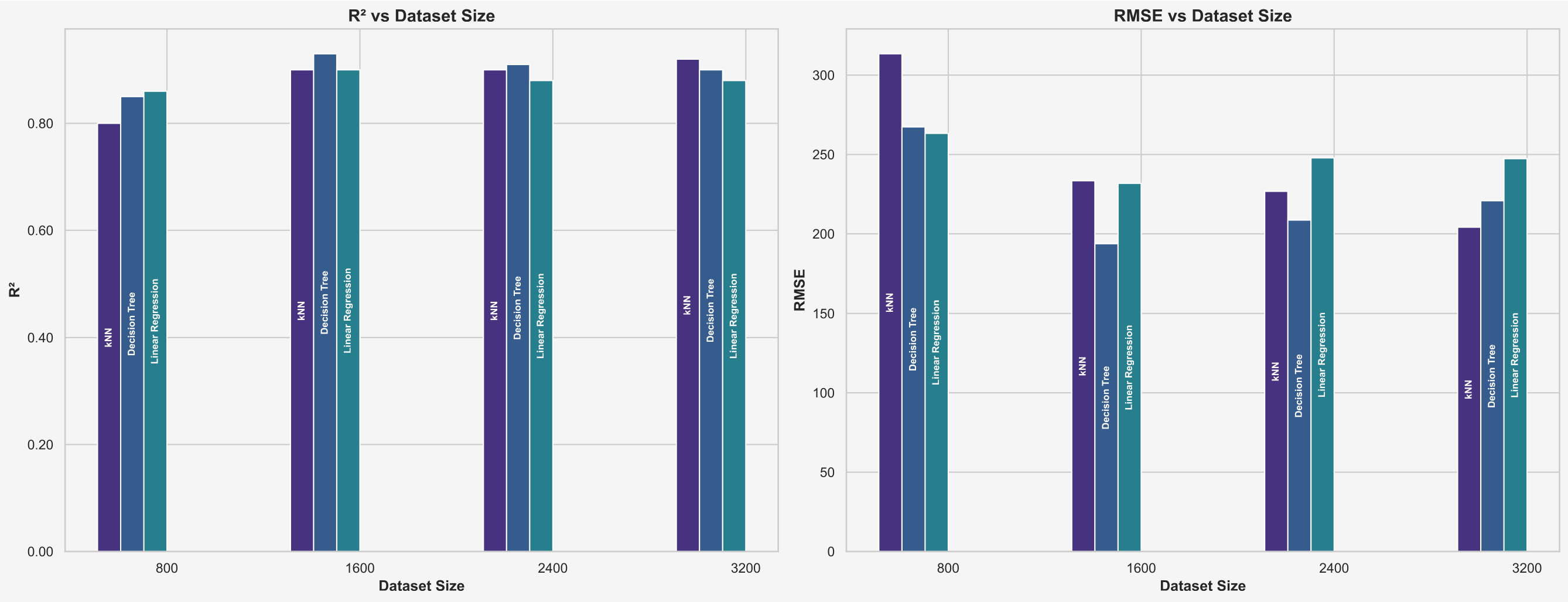}
    \caption{Average $R^2$ and $RMSE$ from cross-validation for classical ML models across different dataset sizes.}
    \label{fig:r2_rmse_bar_plot_classic}
\end{figure}

\begin{table}[h]
    \caption{5-fold cross-validation performance of classical ML models on a dataset of 800 samples.} 
    \label{combined_metrics_800_table_classic}
    \begin{tabular}{@{}lcccccc@{}}
    \toprule
    & \multicolumn{2}{c}{kNN} & \multicolumn{2}{c}{DTR} & \multicolumn{2}{c}{LR} \\
    \cmidrule(lr){2-3} \cmidrule(lr){4-5} \cmidrule(lr){6-7}
    Fold & $R^2$ & RMSE & $R^2$ & RMSE & $R^2$ & RMSE \\
    \mr
    1 & 0.81 & 311.49 & 0.87 & 253.83 & 0.91 & 218.43 \\
    2 & 0.84 & 302.51 & 0.73 & 392.18 & 0.85 & 291.39 \\
    3 & 0.75 & 337.70 & 0.88 & 239.80 & 0.81 & 297.27 \\
    4 & 0.76 & 353.66 & 0.88 & 252.58 & 0.86 & 265.49 \\
    5 & 0.85 & 261.84 & 0.91 & 198.47 & 0.87 & 243.91 \\
    \mr
    \textbf{Avg} & 0.80 $\pm$0.05 & 313.44 $\pm$35.33 & 0.85 $\pm$0.07 & 267.37 $\pm$73.29 & 0.86 $\pm$0.03 & 263.30 $\pm$32.93 \\
    \br
    \end{tabular}
\end{table}
\begin{table}[h]
    \caption{5-fold cross-validation performance of classical ML models on a dataset of 1600 samples.} 
    \label{combined_metrics_1600_table_classic}
    \begin{tabular}{@{}lcccccc@{}}
    \toprule
    & \multicolumn{2}{c}{kNN} & \multicolumn{2}{c}{DTR} & \multicolumn{2}{c}{LR} \\
    \cmidrule(lr){2-3} \cmidrule(lr){4-5} \cmidrule(lr){6-7}
    Fold & $R^2$ & RMSE & $R^2$ & RMSE & $R^2$ & RMSE \\
    \mr
    1 & 0.90 & 234.09 & 0.90 & 230.78 & 0.90 & 235.02 \\
    2 & 0.90 & 228.44 & 0.95 & 172.10 & 0.90 & 236.13 \\
    3 & 0.89 & 232.58 & 0.91 & 210.40 & 0.88 & 238.81 \\
    4 & 0.90 & 225.49 & 0.96 & 142.17 & 0.91 & 211.15 \\
    5 & 0.89 & 246.59 & 0.92 & 213.43 & 0.90 & 238.02 \\
    \mr
    \textbf{Avg} & 0.90 $\pm$0.01 & 233.44 $\pm$8.1 & 0.93 $\pm$0.03 & 193.78 $\pm$35.94 & 0.90 $\pm$0.01 & 231.82 $\pm$11.66 \\
    \br
    \end{tabular}
\end{table}
\begin{table}[h]
    \caption{5-fold cross-validation performance of classical ML models on a dataset of 2400 samples.} 
    \label{combined_metrics_2400_table_classic}
    \begin{tabular}{@{}lcccccc@{}}
    \toprule
    & \multicolumn{2}{c}{kNN} & \multicolumn{2}{c}{DTR} & \multicolumn{2}{c}{LR} \\
    \cmidrule(lr){2-3} \cmidrule(lr){4-5} \cmidrule(lr){6-7}
    Fold & $R^2$ & RMSE & $R^2$ & RMSE & $R^2$ & RMSE \\
    \mr
    1 & 0.90 & 229.64 & 0.94 & 184.28 & 0.90 & 232.83 \\
    2 & 0.88 & 244.08 & 0.88 & 238.41 & 0.83 & 284.03 \\
    3 & 0.89 & 241.45 & 0.88 & 249.42 & 0.88 & 254.76 \\
    4 & 0.90 & 219.25 & 0.90 & 224.33 & 0.86 & 260.73 \\
    5 & 0.92 & 199.50 & 0.96 & 146.87 & 0.92 & 206.92 \\
    \mr
    \textbf{Avg} & 0.90 $\pm$0.02 & 226.78 $\pm$18.19 & 0.91 $\pm$0.03 & 208.66 $\pm$42.44 & 0.88 $\pm$0.03 & 247.85 $\pm$29.25 \\
    \br
    \end{tabular}
\end{table}
\begin{table}[h]
    \caption{5-fold cross-validation performance of classical ML models on a dataset of 3200 samples.} 
    \label{combined_metrics_3200_table_classic}
    \begin{tabular}{@{}lcccccc@{}}
    \toprule
    & \multicolumn{2}{c}{kNN} & \multicolumn{2}{c}{DTR} & \multicolumn{2}{c}{LR} \\
    \cmidrule(lr){2-3} \cmidrule(lr){4-5} \cmidrule(lr){6-7}
    Fold & $R^2$ & RMSE & $R^2$ & RMSE & $R^2$ & RMSE \\
    \mr
    1 & 0.92 & 208.96 & 0.90 & 226.92 & 0.87 & 256.85 \\
    2 & 0.92 & 197.89 & 0.91 & 208.72 & 0.88 & 246.65 \\
    3 & 0.90 & 220.88 & 0.88 & 240.46 & 0.86 & 260.09 \\
    4 & 0.93 & 186.93 & 0.92 & 200.36 & 0.89 & 236.86 \\
    5 & 0.92 & 206.26 & 0.90 & 227.83 & 0.89 & 236.35 \\
    \mr
    \textbf{Avg} & 0.92 $\pm$0.013 & 204.18 $\pm$12.68 & 0.90 $\pm$0.017 & 220.86 $\pm$16.10 & 0.88 $\pm$0.014 & 247.36 $\pm$11.00 \\
    \br
    \end{tabular}
\end{table}
To visualize the ML performance better, we gave the bar chart that shows average $R2$ and average $RMSE$ depending on the dataset size in Figure~\ref{fig:r2_rmse_bar_plot_classic}. According to overall results, kNN emerges as the model that benefits best from the increasing dataset which occurs as decreasing $RMSE$ and also increasing $R^2$.

\subsection{Comparative Analysis}
In this section, we compared QNN configurations with classical ML models. To test each model, we used the hold-out dataset that was created during the data preparation step. Hold-out datasets have the size of 200, 400, 600, 800. We compared them using $R2$ and $RMSE$ as accuracy metrics. Then we visualized predictions of QNN configurations and classical ML models to provide a more graphical approach. Finally, we made an error distribution analysis of each model to assess them.

Table~\ref{tab:comparison_qnn_vs_ml_table} provides quantitative prediction results using $R^2$ and $RMSE$ performance metrics based on dataset size. According to the test results, for all dataset sizes, the quantum approaches consistently provided the lowest $RMSE$. Furthermore, they achieved $R^2$ scores that were better than or equal to those of the classical approaches. These quantitative results show that QNN configurations can compete with classical approaches, particularly for smaller dataset sizes, establishing them as a strong alternative. Among the quantum approaches, QNN-3 shows itself as the model that scales best with increasing dataset size, and have the highest $R^2$ and the lowest $RMSE$ at the largest dataset with an $R^2$ of 0.94, and an $RMSE$ of 174.67 respectively. The best classical model, kNN, achieves an $RMSE$ of 182.99 kW. Compared to this, the best QNN model provides a 4.5\% reduction in error. Although the performance of the classical approaches is initially lower than that of the quantum approaches with a dataset size of 200, it improves as the dataset size increases. Only after the dataset size reaches 600 do some classical models begin to perform better than the several QNN models in terms of $R^2$ and $RMSE$.
  
\begin{figure}
    \centering
    \includegraphics[width=1.1\linewidth]{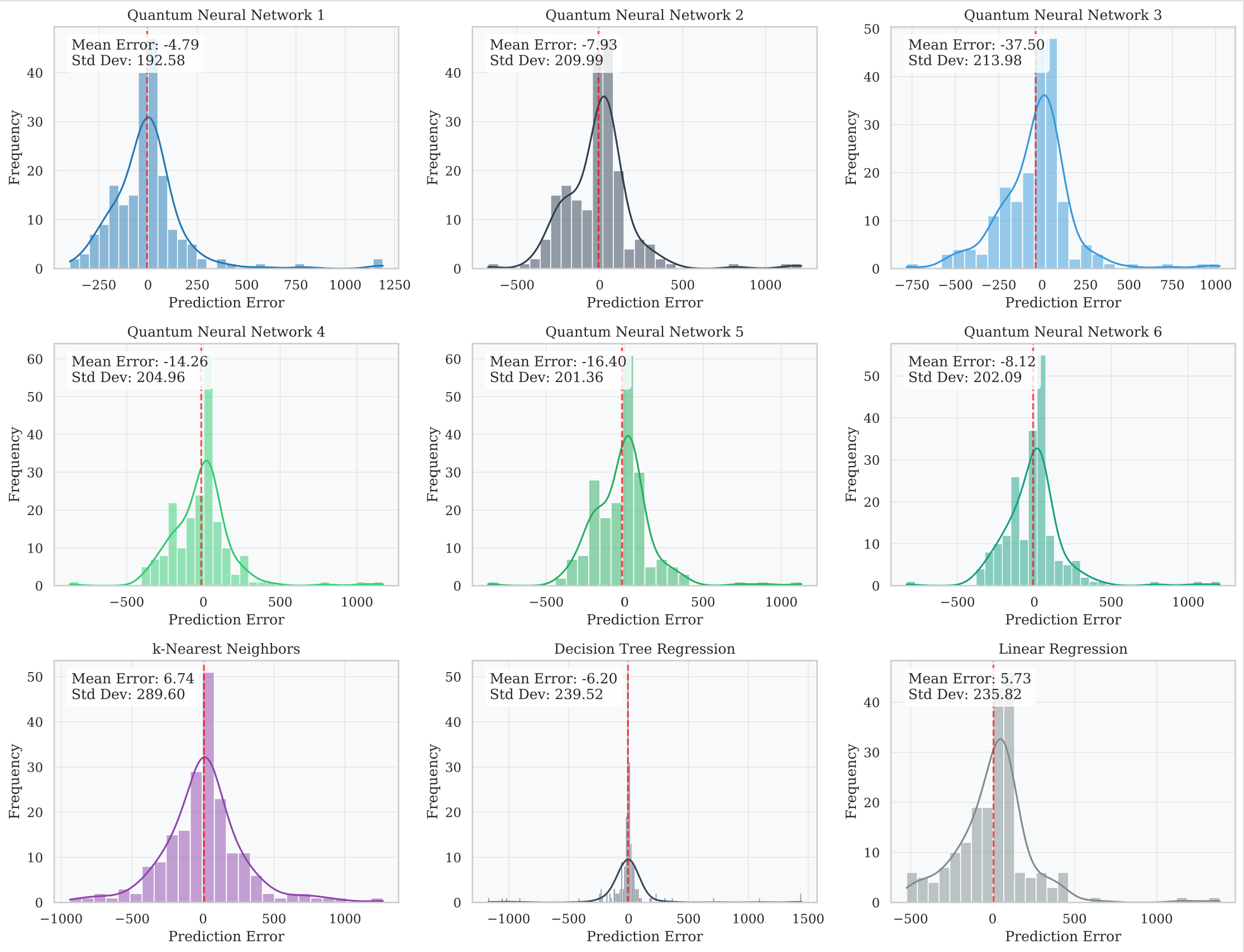}
    \caption{Error distribution plot of actual vs predicted values comparing  QNN and classical models for a test set of 200 samples}
    \label{fig:error_distrib_200}
\end{figure}

\begin{figure}
    \centering
    \includegraphics[width=1.1\linewidth]{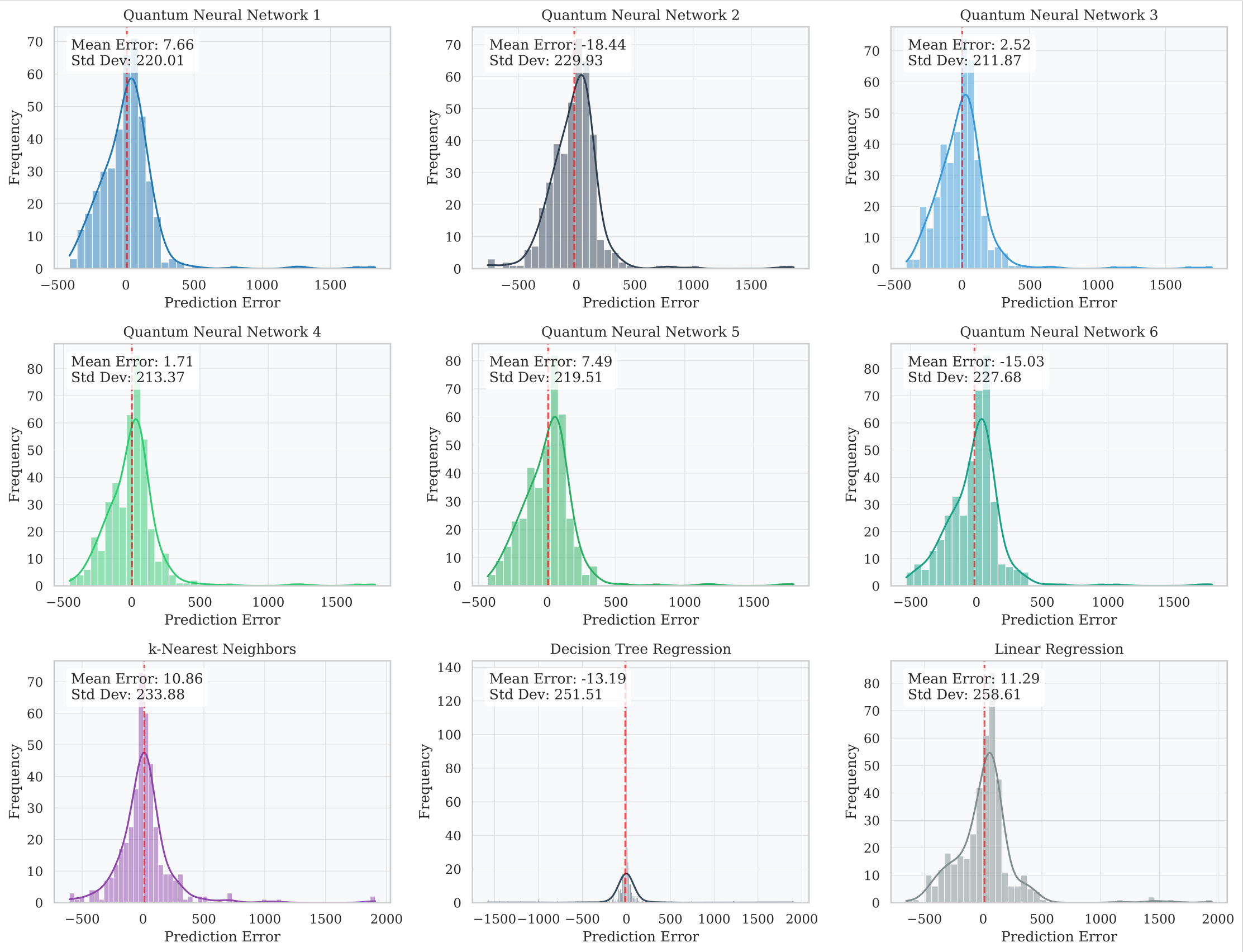}
    \caption{Error distribution plot of actual vs predicted values comparing  QNN and classical models for a test set of 400 samples}
    \label{fig:error_distrib_400}
\end{figure}

\begin{figure}
    \centering
    \includegraphics[width=1.1\linewidth]{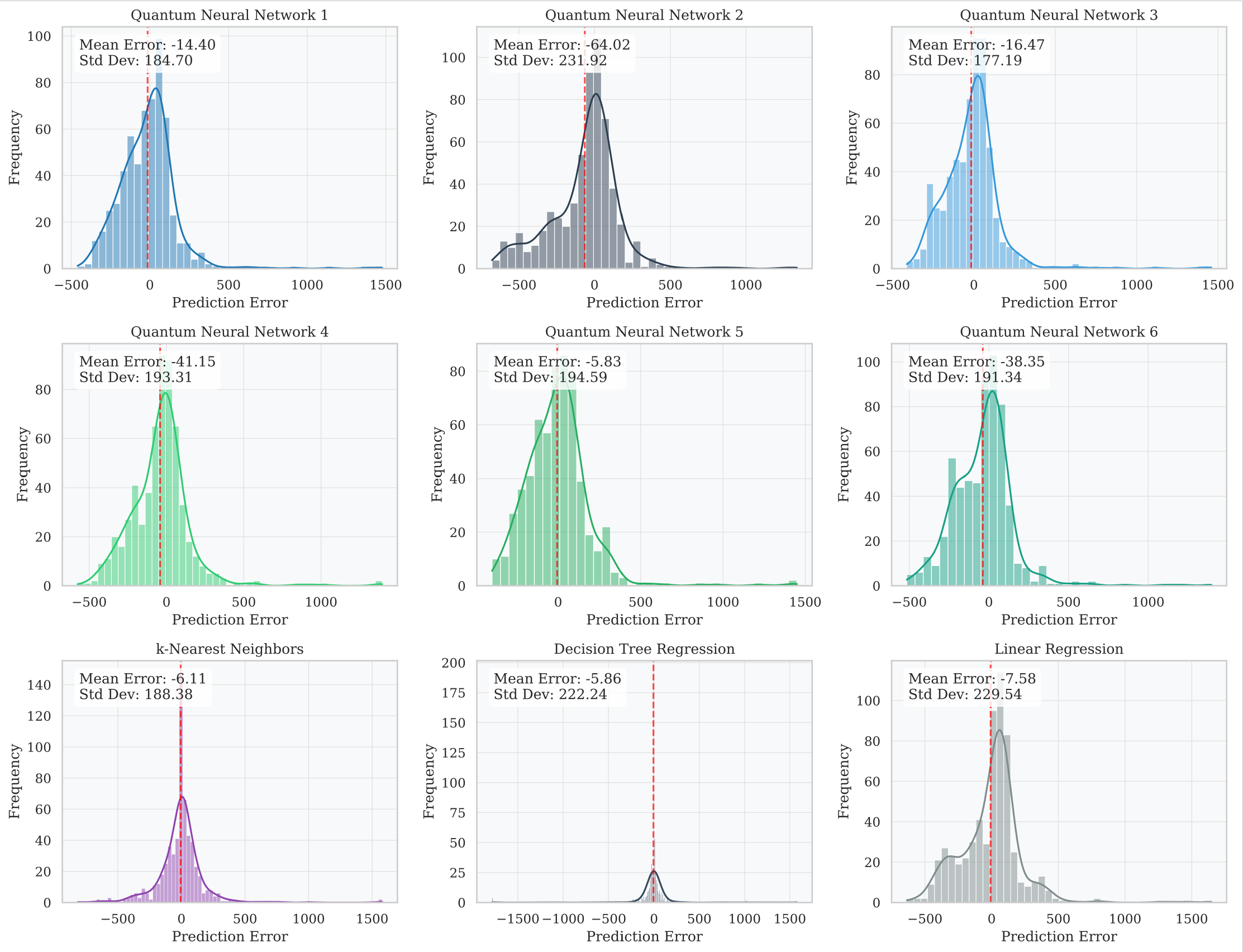}
    \caption{Error distribution plot of actual vs predicted values comparing  QNN and classical models for a test set of 600 samples}
    \label{fig:error_distrib_600}
\end{figure}

\begin{figure}
    \centering
    \includegraphics[width=1.1\linewidth]{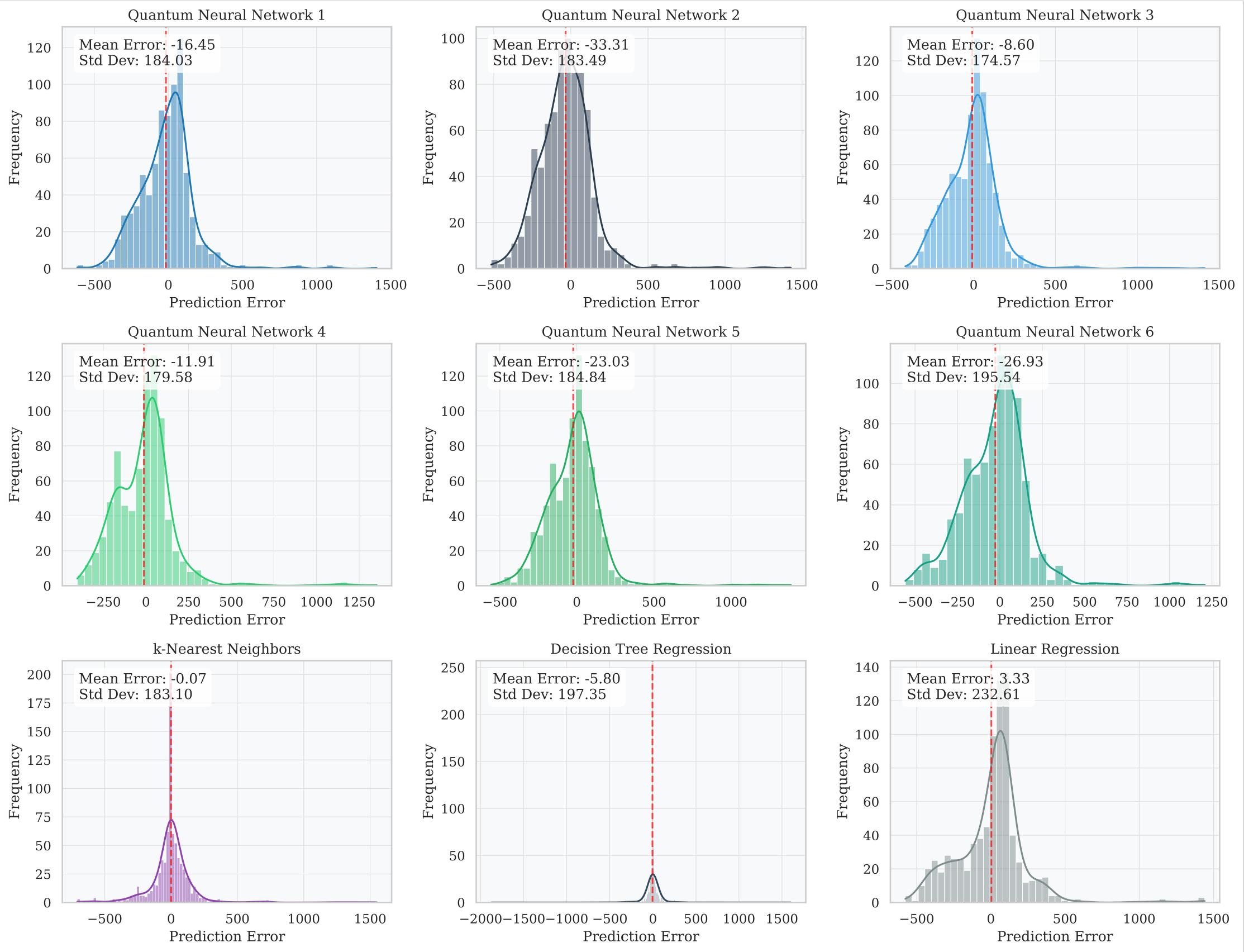}
    \caption{Error distribution plot of actual vs predicted values comparing QNN and classical models for a test set of 800 samples}
    \label{fig:error_distrib_800}
\end{figure}

In the Section~\ref{append2}, we provide scatter plots as Figures~\ref{fig:act_vs_pred_scatter_200},~\ref{fig:act_vs_pred_scatter_400},~\ref{fig:act_vs_pred_scatter_600},~\ref{fig:act_vs_pred_scatter_800} to compare QNN configurations with classical ML to provide a direct visual representation of model performance which can visually support our quantitative results given in Table~\ref{tab:comparison_qnn_vs_ml_table}. Even though scatter plots can visually reflect the model performance metrics shown in Table~\ref{tab:comparison_qnn_vs_ml_table}, for prediction tasks, investigating temporal or sequential consistency is also important. To satisfy that need, we provided Figures~\ref{fig:act_vs_pred_overlay_200},~\ref{fig:act_vs_pred_overlay_400},~\ref{fig:act_vs_pred_overlay_600},~\ref{fig:act_vs_pred_overlay_800}, in the Section~\ref{append3} to give a visual inspection of temporal consistency. According to overlaying plots, although QNN models are good at providing lower error, and higher accuracy, they make fewer mistakes but when they make wrong predictions, we observed negative predicted values, which are not physically possible for a wind turbine. Beside LR, classical models do not have this type of problem. Similar to QNN models, erroneous predictions by LR also contains negative values which are not expected. To address this issue in QNNs, future research should investigate methods for enforcing physical constraints. Potential strategies include the application of an activation function, such as the Rectified Linear Unit (ReLU), or the design of a loss function that penalizes negative values. A thorough evaluation of these approaches is critical for enhancing the real-world applicability of QNNs.
\begin{table}[h] 
\centering
\caption{Performance comparison of QNN and classical models on the hold-out test set across different dataset sizes.} 
\label{tab:comparison_qnn_vs_ml_table}
\begin{tabular}{llcccccccc}
\toprule
\multirow{3}{*}{\textbf{Approach}} & \multirow{3}{*}{} & \multicolumn{8}{c}{\textbf{Dataset size}} \\
\cmidrule(lr){3-10}
& & \multicolumn{2}{c}{\textbf{200}} & \multicolumn{2}{c}{\textbf{400}} & \multicolumn{2}{c}{\textbf{600}} & \multicolumn{2}{c}{\textbf{800}} \\
\cmidrule(lr){3-4} \cmidrule(lr){5-6} \cmidrule(lr){7-8} \cmidrule(lr){9-10}
& & R\textsuperscript{2} & $RMSE$ & R\textsuperscript{2} & $RMSE$ & R\textsuperscript{2} & $RMSE$ & R\textsuperscript{2} & $RMSE$ \\
\mr
\multirow{6}{*}{\textbf{Quantum}} 
& QNN-1 & \textbf{0.92} & \textbf{192.16} & 0.89 & 219.87 & 0.93 & 185.11 & 0.93 & 184.65 \\
& QNN-2 & 0.91 & 209.62 & 0.88 & 230.38 & 0.89 & 240.41 & 0.93 & 186.38 \\
& QNN-3 & 0.90 & 216.71 & \textbf{0.90} & \textbf{211.62} & \textbf{0.94 } & \textbf{177.81} & \textbf{0.94} & \textbf{174.67} \\
& QNN-4 & 0.91 & 204.94 &  0.90 & 213.11 & 0.92 & 197.49 & 0.94   & 179.86 \\
& QNN-5 & 0.92 & 201.53 & 0.89 & 219.36 & 0.93 & 194.51 & 0.93 & 186.16 \\
& QNN-6 & 0.92 & 201.74 & 0.89 & 227.90 & 0.93 & 194.99 & 0.92 & 197.26 \\
\mr
\multirow{3}{*}{\textbf{Classical}} 
& kNN     & 0.83 & 288.96 & 0.88 & 233.84  & 0.93 & 188.32  & 0.94 & 182.99  \\
& DTR    & 0.88 & 239.00 & 0.86 & 251.55 & 0.90 & 222.13 & 0.92 & 197.32 \\
& LR    & 0.89 & 235.30 & 0.85 & 258.54 & 0.90 & 229.47 & 0.90 & 232.49 \\
\br
\end{tabular}
\end{table}
In addition to calculating the $R^2$ and $RMSE$, we also investigated the error distributions for quantum and classical approaches for each dataset. Figures~\ref{fig:error_distrib_200},~\ref{fig:error_distrib_400},~\ref{fig:error_distrib_600},~\ref{fig:error_distrib_800} provides a comparative visualization for error distribution. The quantum approaches exhibited more symmetric and concentrated error distributions than classical models as it was seen at Figures~\ref{fig:error_distrib_200},~\ref{fig:error_distrib_400},~\ref{fig:error_distrib_600},~\ref{fig:error_distrib_800}. For the dataset with least samples (200), QNN-1 and QNN-5 performed best with the tight error spread with standard deviation of  $201.36$ and $192.58$, respectively, and the lowest bias with mean error of $-1.64$ and $-4.79$, respectively. With the dataset size increased from 200 to 400 and 600 samples, QNN-3 emerged as a candidate with best stability that has the standard deviations of $211.87$ and $177.19$. At the largest dataset with 800 samples, QNN-1 and QNN-3 developed a superior predictive stability with the standard deviations of $174.57$ and $184.03$. From the classical models, kNN emerges as a candidate with nearly zero bias. At Table~\ref{tab:holdout_error_rank} gives a summary of the top performers for each dataset based on bias and variance in prediction errors.

\begin{table}[ht]
\centering
\caption{Model ranking by prediction error distribution. The evaluation considers both mean error (bias) and standard deviation (variance).}
\label{tab:holdout_error_rank}
\begin{tabular}{llll}
\toprule
\textbf{Dataset Size} & \textbf{1st Rank} & \textbf{2nd Rank} & \textbf{3rd Rank} \\
\mr
200 & QNN-5 & QNN-1 & QNN-6 \\
400 & QNN-4 & QNN-3 & QNN-5 \\
600 & QNN-3 & QNN-5 & QNN-1 \\
800 & QNN-3 & QNN-1 & kNN \\
\br
\end{tabular}
\end{table}

\subsection{Summary of Experimental Findings}
Our in-depth experimental results demonstrate a detailed assessment of six QNN configurations that have varying ansatz circuit on machine learning for real-world regression tasks. We evaluated QNN configurations regarding four distinct dataset size (800, 1600, 2400, 3200) using the k-fold cross-validation test where $k=5$. According to analytical results, QNN configurations achieved significant prediction accuracy with $R^2$ up to 0.94. Analysis on the cross-validation step revealed that peak performance for QNNs was achieved with the dataset of 1600 samples; further increases in data did not result in performance gains.In contrast to the widely held expectation of performance gains with larger datasets, our findings indicate that an optimal data-to-model-complexity ratio may exist for these ansatz configurations. This suggests a need for future studies on the expressive capacity of different ansätze as a function of dataset size, in order to establish a framework for preventing model overfitting. Investigation of loss function confirmed the rapid convergence for all QNN configurations. Although the loss function typically converges within approximately 10 iterations, increased dataset size contributed to a more stable training process. Overall analysis showed that QNN-6 is the most stable configuration. Finally, we investigated simulation time performance for all configurations. According to results, simulation time performance scaled linearly with dataset size, provides a computational complexity of $\mathcal{O}(n)$. The total number of gates also proved to be a key parameter influencing time performance. The time performance evaluation showed that QNN-5 and QNN-6 are the most time-efficient models.

After the assessment of QNN configurations, we compared quantum approach with classical approaches such as kNN, DTR, and LR. For comparison, we used hold-out test datasets derived from the original dataset. Especially for smaller datasets, quantum models demonstrated superior machine learning performance, in particular for smaller datasets. In detail, QNN-3 and QNN-4, achieved lower $RMSE$, and higher $R^2$ compared to their classical counterparts. For the largest hold-out dataset (800 samples), QNN-3 yielded an $R^2$ of 0.94, and $RMSE$ of 174.67 kW, which can be significiantly comparable with the best classical model, kNN which achieved an $R^2$ of 0.94 and $RMSE$ of 182.99 kW. Experimental findings quantitatively showcased the potential of QNNs to offer significant prediction accuracy and robustness, showing them as a competitive and promising alternative to classical approaches when it comes to solving complex real-world prediction problems.

\section{Conclusion and Future Work}\label{conclusion}
In this study, we conducted a systematic and comprehensive benchmark of six QNN configurations against three classical approaches such as kNN, DTR, and LR. We selected a real-world application that is related to wind turbine power forecasting. We built varying quantum circuits based on Z Feature Map and different ansatz structures. Then we investigated the impact of dataset size and quantum circuit complexity on both simulation time efficiency and machine learning performance. Our in-depth experiments demonstrate that QNNs represent a promising alternative to classical models, demonstrating particular strength in certain data-limited scenarios explored in this study. Our key findings revealed that, QNN configurations achieved high accuracy, with an $R^2$ score up to 0.94, and competed with classical approaches on both training and test datasets. QNN-3 emerged as the best-performing QNN model and provided comparable results to all of the classical models. 

By conducting a cross-validation analysis, we found that the peak performance of the QNN configurations was obtained with the dataset containing 1600 samples. Further increases in the dataset size did not yield additional performance gains. This observation suggests an optimal data-to-model-complexity ratio may exist for this regression problem, which contradicts the common assumption that larger datasets always improve performance. We also observed that while QNNs provided superior results in terms of $R^2$ and $RMSE$, they occasionally produced predictions outside the target feature's physical scope, such as negative power values. This observation suggests that the pre-processing of data in the classical domain might hold importance and is worth investigating. 

Since we are in the NISQ era of quantum computing and access to real quantum hardware is limited, classical hardware-based simulations remain important. To provide insight about simulation time performance, we investigated the time performance of QNN models on a linearly scaling dataset. Based on experiments, quantum approach has a time complexity of $\mathcal{O}(n)$. Additionally, time performance was directly correlated with circuit complexity, as determined by the total gate count. Independent of gate sequence or gate connections, quantum circuits that have the same number of total gates yielded similar time performance. Circuits that have Reverse Linear ansatz (QNN-5) and Pairwise ansatz (QNN-6) proved to be the most time-efficient of QNN configurations. 

Regarding the stability of the learning phase, we also investigated the training dynamics and stability of the QNN models. According to results, all QNN models demonstrated rapid convergence, and stability within the first 10-15 iterations of the training step. Results revealed that the larger datasets contributed to more stable training dynamics and lower final loss values. QNN-6 proved to be the most stable model across all dataset sizes. It is also worth noting that due to the significant time required to train the QNN configurations, hyper-parameter tuning was left outside the scope of this study. To ensure a fair comparison, the classical models were also not tuned and were run with their default parameters. In future work, we will subject the best-performing models from this study to a hyper-parameter tuning benchmark, utilizing hardware-accelerated simulation libraries to determine if comparable or improved results can be achieved with reduced training times. 

Another important point is, we compared QNN configurations with several classical approaches that have been proven to be successful in our previous study. To provide better insight, QNN configurations might need to be compared with other classical approaches that are commonly used for prediction tasks on time-series data. Gradient Boosting Machines (e.g., XGBoost, LightGBM), Support Vector Regression (SVR) are common approaches that are used for such tasks. So, for future work, a comparison between QNN configurations and these methods is also important to prove QNNs' potential.

In conclusion, this research provides practical insights for applying QML based on QNNs to real-world regression tasks. Results highlight that chosen QNN configurations can provide a significant advantage over classical models in predictive accuracy, particularly in data-limited scenarios. Our findings serve as an experimental guide for researchers helping to observe the trade-offs between stability, model performance, and the computational costs associated with quantum machine learning approaches.

\section*{Acknowledgements}
The authors would like to acknowledge that this paper is submitted in partial fulfillment of the requirements for the PhD degree at Yildiz Technical University. The Open Access publication fee of this article was supported by TÜBİTAK (The Scientific and Technological Research Council of Türkiye) under its Open Access Support Program.

\section*{Declarations}
For this research paper, no funds, grants, or other support were received. All authors certify that they have no affiliations with or involvement in any organization or entity with any financial interest or non-financial interest in the subject matter or materials discussed in this manuscript.

\section*{Declaration of generative AI and AI-assisted technologies in the
writing process}
During the preparation of this work, the authors used Google Gemini to assist with minor language editing. After using this tool, the authors reviewed and edited the content as necessary and take full responsibility for the content of the publication.

\section*{References}
\bibliography{main.bib}

\providecommand{\newblock}{}
\begin{thebibliography}{10}
\expandafter\ifx\csname url\endcsname\relax
  \def\url#1{{\tt #1}}\fi
\expandafter\ifx\csname urlprefix\endcsname\relax\def\urlprefix{URL }\fi
\providecommand{\eprint}[2][]{\url{#2}}

\bibitem{willow2024}
Acharya R, Abanin D~A, Aghababaie-Beni L, Aleiner I, Andersen T~I, Ansmann M,
  Arute F, Arya K, Asfaw A, Astrakhantsev N, Atalaya J, Babbush R, Bacon D,
  Ballard B, Bardin J~C, Bausch J, Bengtsson A, Bilmes A, Blackwell S, Boixo S,
  Bortoli G, Bourassa A, Bovaird J, Brill L, Broughton M, Browne D~A, Buchea B,
  Buckley B~B, Buell D~A, Burger T, Burkett B, Bushnell N, Cabrera A, Campero
  J, Chang H~S, Chen Y, Chen Z, Chiaro B, Chik D, Chou C, Claes J, Cleland A~Y,
  Cogan J, Collins R, Conner P, Courtney W, Crook A~L, Curtin B, Das S, Davies
  A, De~Lorenzo L, Debroy D~M, Demura S, Devoret M, Di~Paolo A, Donohoe P,
  Drozdov I, Dunsworth A, Earle C, Edlich T, Eickbusch A, Elbag A~M, Elzouka M,
  Erickson C, Faoro L, Farhi E, Ferreira V~S, Burgos L~F, Forati E, Fowler A~G,
  Foxen B, Ganjam S, Garcia G, Gasca R, Genois E, Giang W, Gidney C, Gilboa D,
  Gosula R, Dau A~G, Graumann D, Greene A, Gross J~A, Habegger S, Hall J,
  Hamilton M~C, Hansen M, Harrigan M~P, Harrington S~D, Heras F~J~H, Heslin S,
  Heu P, Higgott O, Hill G, Hilton J, Holland G, Hong S, Huang H~Y, Huff A,
  Huggins W~J, Ioffe L~B, Isakov S~V, Iveland J, Jeffrey E, Jiang Z, Jones C,
  Jordan S, Joshi C, Juhas P, Kafri D, Kang H, Karamlou A~H, Kechedzhi K, Kelly
  J, Khaire T, Khattar T, Khezri M, Kim S, Klimov P~V, Klots A~R, Kobrin B,
  Kohli P, Korotkov A~N, Kostritsa F, Kothari R, Kozlovskii B, Kreikebaum J~M,
  Kurilovich V~D, Lacroix N, Landhuis D, Lange-Dei T, Langley B~W, Laptev P,
  Lau K~M, Le~Guevel L, Ledford J, Lee J, Lee K, Lensky Y~D, Leon S, Lester
  B~J, Li W~Y, Li Y, Lill A~T, Liu W, Livingston W~P, Locharla A, Lucero E,
  Lundahl D, Lunt A, Madhuk S, Malone F~D, Maloney A, Mandra S, Manyika J,
  Martin L~S, Martin O, Martin S, Maxfield C, McClean J~R, McEwen M, Meeks S,
  Megrant A, Mi X, Miao K~C, Mieszala A, Molavi R, Molina S, Montazeri S,
  Morvan A, Movassagh R, Mruczkiewicz W, Naaman O, Neeley M, Neill C, Nersisyan
  A, Neven H, Newman M, Ng J~H, Nguyen A, Nguyen M, Ni C~H, Niu M~Y, O'Brien
  T~E, Oliver W~D, Opremcak A, Ottosson K, Petukhov A, Pizzuto A, Platt J,
  Potter R, Pritchard O, Pryadko L~P, Quintana C, Ramachandran G, Reagor M~J,
  Redding J, Rhodes D~M, Roberts G, Rosenberg E, Rosenfeld E, Roushan P, Rubin
  N~C, Saei N, Sank D, Sankaragomathi K, Satzinger K~J, Schurkus H~F, Schuster
  C, Senior A~W, Shearn M~J, Shorter A, Shutty N, Shvarts V, Singh S, Sivak V,
  Skruzny J, Small S, Smelyanskiy V, Smith W~C, Somma R~D, Springer S, Sterling
  G, Strain D, Suchard J, Szasz A, Sztein A, Thor D, Torres A, Torunbalci M~M,
  Vaishnav A, Vargas J, Vdovichev S, Vidal G, Villalonga B, Heidweiller C~V,
  Waltman S, Wang S~X, Ware B, Weber K, Weidel T, White T, Wong K, Woo B~W~K,
  Xing C, Yao Z~J, Yeh P, Ying B, Yoo J, Yosri N, Young G, Zalcman A, Zhang Y,
  Zhu N, Zobrist N and Collaborato G~Q~A 2024 {\em NATURE\/} ISSN 0028-0836

\bibitem{majorona2025}
Aghaee M, Ramirez A~A, Alam Z, Ali R, Andrzejczuk M, Antipov A, Astafev M,
  Barzegar A, Bauer B, Becker J, Bhaskar U~K, Bocharov A, Boddapati S, Bohn D,
  Bommer J, Bourdet L, Bousquet A, Boutin S, Casparis L, Chapman B~J, Chatoor
  S, Christensen A~W, Chua C, Codd P, Cole W, Cooper P, Corsetti F, Cui A,
  Dalpasso P, Dehollain J~P, de~Lange G, de~Moor M, Ekefjard A, El~Dandachi T,
  Saldana J~C~E, Fallahi S, Galletti L, Gardner G, Govender D, Griggio F,
  Grigoryan R, Grijalva S, Gronin S, Gukelberger J, Hamdast M, Hamze F, Hansen
  E~B, Heedt S, Heidarnia Z, Zamorano J~H, Ho S, Holgaard L, Hornibrook J,
  Indrapiromkul J, Ingerslev H, Ivancevic L, Jensen T, Jhoja J, Jones J,
  Kalashnikov K~V, Kallaher R, Kalra R, Karimi F, Karzig T, King E, Kloster
  M~E, Knapp C, Kocon D, Koski J~V, Kostamo P, Kumar M, Laeven T, Larsen T, Lee
  J, Lee K, Leum G, Li K, Lindemann T, Looij M, Love J, Lucas M, Lutchyn R,
  Madsen M~H, Madulid N, Malmros A, Manfra M, Mantri D, Markussen S~B, Martinez
  E, Mattila M, McNeil R, Mei A~B, Mishmash R~V, Mohandas G, Mollgaard C,
  Morgan T, Moussa G, Nayak C, Nielsen J~H, Nielsen J~M, Nielsen W~H~P, Nijholt
  B, Nystrom M, O'Farrell E, Ohki T, Otani K, Wutz B~P, Pauka S, Petersson K,
  Petit L, Pikulin D, Prawiroatmodjo G, Preiss F, Morejon E~P, Rajpalke M,
  Ranta C, Rasmussen K, Razmadze D, Reentila O, Reilly D~J, Ren Y, Reneris K,
  Rouse R, Sadovskyy I, Sainiemi L, Sanlorenzo I, Schmidgall E, Sfiligoj C,
  Shah M~B, Simoes K, Singh S, Sinha S, Soerensen T, Sohr P, Stankevic T, Stek
  L, Stuppard E, Suominen H, Suter J, Teicher S, Thiyagarajah N, Tholapi R,
  Thomas M, Toomey E, Tracy J, Turley M, Upadhyay S, Urban I, Van~Hoogdalem K,
  Van~Woerkom D~J, Viazmitinov D~V, Vogel D, Watson J, Webster A, Weston J,
  Winkler G~W, Xu D, Yang C~K, Yucelen E, Zeisel R, Zheng G, Zilke J and
  Quantum M~A 2025 {\em NATURE\/} {\bf 638} 651+ ISSN 0028-0836

\bibitem{TomSvore2024}
Tom D and Svore K 2024 How microsoft and quantinuum achieved reliable quantum
  computing microsoft Azure Quantum Blog
  \urlprefix\url{https://azure.microsoft.com/en-us/blog/quantum/2024/04/03/how-microsoft-and-quantinuum-achieved-reliable-quantum-computing/}

\bibitem{microsoft2024}
Paetznick A, da~Silva M~P, Ryan-Anderson C, Bello-Rivas J~M, III J~P~C,
  Chernoguzov A, Dreiling J~M, Foltz C, Frachon F, Gaebler J~P, Gatterman T~M,
  Grans-Samuelsson L, Gresh D, Hayes D, Hewitt N, Holliman C, Horst C~V,
  Johansen J, Lucchetti D, Matsuoka Y, Mills M, Moses S~A, Neyenhuis B, Paz A,
  Pino J, Siegfried P, Sundaram A, Tom D, Wernli S~J, Zanner M, Stutz R~P and
  Svore K~M 2024 Demonstration of logical qubits and repeated error correction
  with better-than-physical error rates (\textit{Preprint} \eprint{2404.02280})
  \urlprefix\url{https://arxiv.org/abs/2404.02280}

\bibitem{kyriienko2022}
Kyriienko O and Magnusson E~B 2022 {\em arXiv preprint arXiv:2208.01203\/}

\bibitem{sakhnenko2022hybrid}
Sakhnenko A, O’Meara C, Ghosh K~J, Mendl C~B, Cortiana G and
  Bernab{\'e}-Moreno J 2022 {\em Quantum Machine Intelligence\/} {\bf 4} 27

\bibitem{javaria2023}
Amin J, Anjum M~A, Ibrar K, Sharif M, Kadry S and Crespo R~G 2023 {\em Image
  and Vision Computing\/} {\bf 135} 104710 ISSN 0262-8856
  \urlprefix\url{https://www.sciencedirect.com/science/article/pii/S0262885623000847}

\bibitem{wang2023}
Wang M, Huang A, Liu Y, Yi X, Wu J and Wang S 2023 {\em Entropy\/} {\bf 25}
  ISSN 1099-4300 \urlprefix\url{https://www.mdpi.com/1099-4300/25/3/427}

\bibitem{sha2024}
Sha M and Rahamathulla M~P 2024 {\em Quantum Information Processing\/} {\bf 23}
  ISSN 1570-0755

\bibitem{zhuang2024}
Zhuang S, Tanner J, Wu Y, Huynh D, Liu W, Cadet X, Fontaine N, Charton P,
  Damour C, Cadet F and Wang J 2024 {\em Quantum Information Processing\/} {\bf
  23} ISSN 1570-0755

\bibitem{cheny2024}
Chen Y 2024 {\em Quantum Information Processing\/} {\bf 23} ISSN 1570-0755

\bibitem{abdulsalam2025}
Abdulsalam G and Ahmad I 2025 {\em Quantum Information Processing\/} {\bf 24}
  109 ISSN 1573-1332 \urlprefix\url{https://doi.org/10.1007/s11128-025-04728-3}

\bibitem{yang2025}
Yang Q, Zhang W and Wei L 2025 {\em Quantum Information Processing\/} {\bf 24}
  ISSN 1570-0755

\bibitem{ajagekar2021}
Ajagekar A and You F 2021 {\em Applied Energy\/} {\bf 303} 117628 ISSN
  0306-2619

\bibitem{zhou2022noise}
Zhou Y and Zhang P 2023 {\em IEEE Transactions on Power Systems\/} {\bf 38}
  475--487

\bibitem{yu2023prediction}
Yu Y, Hu G, Liu C, Xiong J and Wu Z 2023 {\em IEEE Transactions on Quantum
  Engineering\/} {\bf 4} 1--15

\bibitem{satpathy2024}
Satpathy S~K, Vibhu V, Behera B~K, Al-Kuwari S, Mumtaz S and Farouk A 2024 {\em
  IEEE Internet of Things Journal\/} {\bf 11} 3840--3852

\bibitem{chen2024}
Chen J and Li Y 2024 Extended abstract: Quantum-accelerated transient stability
  assessment for power systems {\em 2024 IEEE Computer Society Annual Symposium
  on VLSI (ISVLSI)\/} pp 593--594

\bibitem{safari2024neuroquman}
Safari A and Badamchizadeh M~A 2024 {\em Neural Computing and Applications\/}
  {\bf 36} 19121--19138

\bibitem{hangun2024SmartGrid}
Hangun B, Eyecioglu O and Altun O 2024 Quantum computing approach to smart grid
  stability forecasting {\em 2024 12th International Conference on Smart Grid
  (icSmartGrid)\/} pp 840--843

\bibitem{hangun2024wind}
Hangun B, Akpinar E, Oduncuoglu M, Altun O and Eyecioglu O 2024 A hybrid
  quantum-classical machine learning approach to offshore wind farm power
  forecasting {\em 2024 13th International Conference on Renewable Energy
  Research and Applications (ICRERA)\/} pp 1105--1110

\bibitem{gwec2024}
{Global Wind Energy Council} 2024 Global wind report 2024
  \url{https://gwec.net/global-wind-report-2024/} accessed: 2025-04-09
  \urlprefix\url{https://gwec.net/global-wind-report-2024/}

\bibitem{gregor2011}
Giebel G, Brownsword R, Kariniotakis G, Denhard M and Draxl C 2011 {\em The
  State-Of-The-Art in Short-Term Prediction of Wind Power: A Literature
  Overview, 2nd edition\/} (ANEMOS.plus) project funded by the European
  Commission under the 6th Framework Program, Priority 6.1: Sustainable Energy
  Systems

\bibitem{foley2012}
Foley A~M, Leahy P~G, Marvuglia A and McKeogh E~J 2012 {\em Renewable Energy\/}
  {\bf 37} 1--8 ISSN 0960-1481

\bibitem{ju2019}
Ju Y, Sun G, Chen Q, Zhang M, Zhu H and Rehman M~U 2019 {\em IEEE Access\/}
  {\bf 7} 28309--28318

\bibitem{shabbir2019}
Shabbir N, AhmadiAhangar R, Kütt L, Iqbal M~N and Rosin A 2019 Forecasting
  short term wind energy generation using machine learning {\em 2019 IEEE 60th
  International Scientific Conference on Power and Electrical Engineering of
  Riga Technical University (RTUCON)\/} pp 1--4

\bibitem{sulaiman2024}
Sulaiman M~H and Mustaffa Z 2024 {\em Cleaner Energy Systems\/} {\bf 9} 100139
  ISSN 2772-7831

\bibitem{olcay2024}
Olcay K, Gíray~Tunca S and Aríf~Özgür M 2024 {\em IEEE Access\/} {\bf 12}
  103299--103312

\bibitem{ajagekar2019}
Ajagekar A and You F 2019 {\em Energy\/} {\bf 179} 76--89 ISSN 0360-5442

\bibitem{zhou2022}
Zhou Y, Tang Z, Nikmehr N, Babahajiani P, Feng F, Wei T~C, Zheng H and Zhang P
  2022 {\em iEnergy\/} {\bf 1} 170--187

\bibitem{zhou2023}
Zhou Y and Zhang P 2023 {\em IEEE Transactions on Power Systems\/} {\bf 38}
  475--487

\bibitem{ranga2024}
Ranga D, Rana A, Prajapat S, Kumar P, Kumar K and Vasilakos A~V 2024 {\em
  Mathematics\/} {\bf 12} ISSN 2227-7390
  \urlprefix\url{https://www.mdpi.com/2227-7390/12/21/3318}

\bibitem{hong2025}
Hong Y~Y and Santos J~B~D 2025 {\em Energies\/} {\bf 18} ISSN 1996-1073
  \urlprefix\url{https://www.mdpi.com/1996-1073/18/7/1771}

\bibitem{gujju2024}
Gujju Y, Matsuo A and Raymond R 2024 {\em Phys. Rev. Appl.\/} {\bf 21}(6)
  067001
  \urlprefix\url{https://link.aps.org/doi/10.1103/PhysRevApplied.21.067001}

\bibitem{budinski2023}
Budinski L, Niemimäki O, Zamora-Zamora R and Lahtinen V 2023 {\em Quantum
  Science and Technology\/} {\bf 8} 045031
  \urlprefix\url{https://dx.doi.org/10.1088/2058-9565/acfab7}

\bibitem{smith2023}
Smith A~W~R, Paige A~J and Kim M~S 2023 {\em Quantum Science and Technology\/}
  {\bf 8} 045016 \urlprefix\url{https://dx.doi.org/10.1088/2058-9565/aceb87}

\bibitem{mineh2023}
Mineh L and Montanaro A 2023 {\em Quantum Science and Technology\/} {\bf 8}
  035012 \urlprefix\url{https://dx.doi.org/10.1088/2058-9565/acd0d2}

\bibitem{agliardi2025}
Agliardi G, O’Meara C, Yogaraj K, Ghosh K, Sabino P, Fernández-Campoamor M,
  Cortiana G, Bernabé-Moreno J, Tacchino F, Mezzacapo A and Shehab O 2025 {\em
  Quantum Science and Technology\/} {\bf 10} 025005
  \urlprefix\url{https://dx.doi.org/10.1088/2058-9565/ada08c}

\bibitem{ponce2025}
Ponce M, Cope T, de~Vega I and Leib M 2024 {\em Quantum Science and
  Technology\/} {\bf 10} 015027
  \urlprefix\url{https://dx.doi.org/10.1088/2058-9565/ad8eee}

\bibitem{hangunWind2025}
Hangun B, Akpinar E, Altun O and Eyecioglu O 2025 Comparative analysis of qnn
  architectures for wind power prediction: Feature maps and ansatz
  configurations (\textit{Preprint} \eprint{2506.14795})
  \urlprefix\url{https://arxiv.org/abs/2506.14795}

\bibitem{eyecioglu2019}
Eyecioglu O, Hangun B, Kayisli K and Yesilbudak M 2019 Performance comparison
  of different machine learning algorithms on the prediction of wind turbine
  power generation {\em 2019 8th International Conference on Renewable Energy
  Research and Applications (ICRERA)\/} pp 922--926

\bibitem{DTU_Risoe_WindDatabase}
of~Denmark T~U and Laboratory R~N Database on wind characteristics
  \url{http://www.winddata.com} [Online]

\bibitem{Schuld2015}
Schuld M, Sinayskiy I and and F~P 2015 {\em Contemporary Physics\/} {\bf 56}
  172--185 (\textit{Preprint}
  \eprint{https://doi.org/10.1080/00107514.2014.964942})
  \urlprefix\url{https://doi.org/10.1080/00107514.2014.964942}

\bibitem{preskill2015}
Preskill J 2018 {\em {Quantum}\/} {\bf 2} 79 ISSN 2521-327X
  \urlprefix\url{https://doi.org/10.22331/q-2018-08-06-79}

\bibitem{rath2024quantum}
Rath M and Date H 2024 {\em EPJ Quantum Technology\/} {\bf 11} 72

\bibitem{yuxuan2022}
Du Y, Huang T, You S, Hsieh M~H and Tao D 2022 {\em NPJ QUANTUM INFORMATION\/}
  {\bf 8}

\bibitem{schuld2019}
Schuld M and Killoran N 2019 {\em Physical Review Letters\/} {\bf 122} 040504

\bibitem{qiskit2024}
Javadi-Abhari A, Treinish M, Krsulich K, Wood C~J, Lishman J, Gacon J, Martiel
  S, Nation P~D, Bishop L~S, Cross A~W, Johnson B~R and Gambetta J~M 2024
  Quantum computing with {Q}iskit (\textit{Preprint} \eprint{2405.08810})

\bibitem{havlicek2019}
Havl{\'\i}{\v{c}}ek V, C{\'o}rcoles A~D, Temme K, Harrow A~W, Kandala A, Chow
  J~M and Gambetta J~M 2019 {\em Nature\/} {\bf 567} 209--212

\bibitem{mcclean2016}
McClean J~R, Romero J, Babbush R and Aspuru-Guzik A 2016 {\em NEW JOURNAL OF
  PHYSICS\/} {\bf 18} ISSN 1367-2630

\bibitem{abbas2021}
Abbas A, Sutter D, Zoufal C, Lucchi A, Figalli A and Woerner S 2021 {\em NATURE
  COMPUTATIONAL SCIENCE\/} {\bf 1} 403--409

\bibitem{arthur2022_hybrid}
Arthur D and Date P 2022 A hybrid quantum-classical neural network architecture
  for binary classification (\textit{Preprint} \eprint{2201.01820})
  \urlprefix\url{https://arxiv.org/abs/2201.01820}

\bibitem{qiskitml2025}
Sahin M~E, Altamura E, Wallis O, Wood S~P, Dekusar A, Millar D~A, Imamichi T,
  Matsuo A and Mensa S 2025 Qiskit machine learning: an open-source library for
  quantum machine learning tasks at scale on quantum hardware and classical
  simulators (\textit{Preprint} \eprint{2505.17756})
  \urlprefix\url{https://arxiv.org/abs/2505.17756}

\end{thebibliography}


\clearpage

\appendix

\clearpage
\section{Quantum circuits used in this study}\label{append1}

\begin{figure}[h!]
    \centering
    \includegraphics[width=0.45\linewidth]{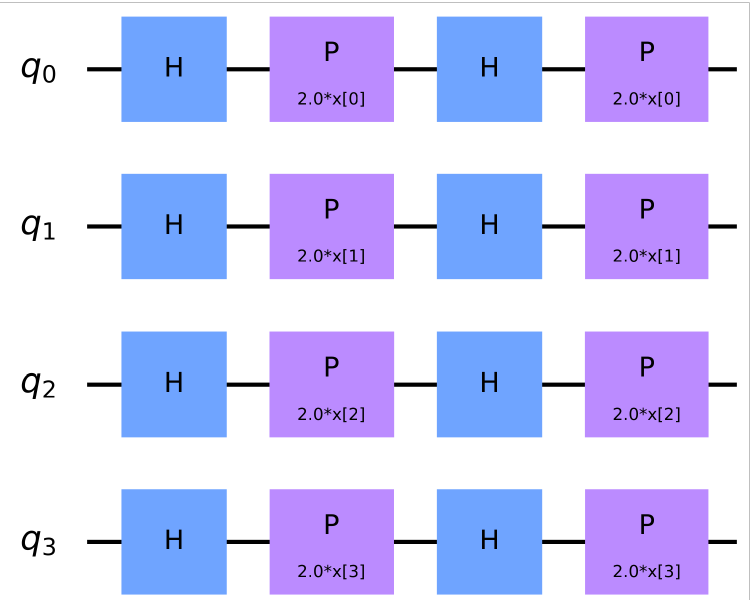}
    \caption{Quantum circuit for the Z Feature Map with two repetitions.}
    \label{fig:appendix_z_feature_map_circ}
\end{figure}

\begin{figure}[h!]
    \centering
    \includegraphics[width=0.65\linewidth]{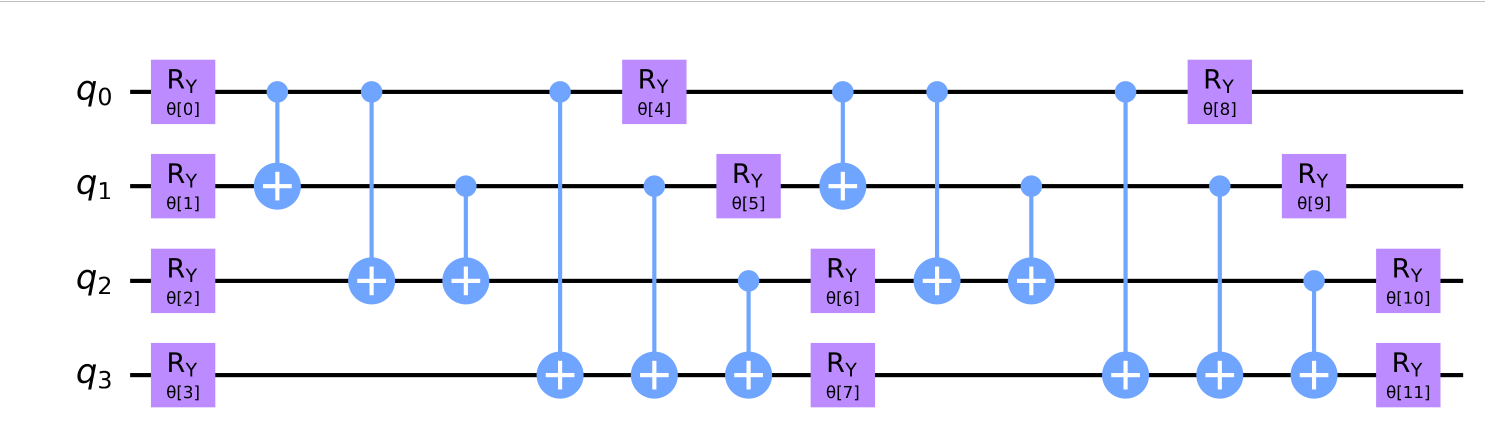}
    \caption{Ansatz with full entanglement.}
    \label{fig:full_ansatz_circ}
\end{figure}

\begin{figure}[h!]
    \centering
    \includegraphics[width=0.85\linewidth]{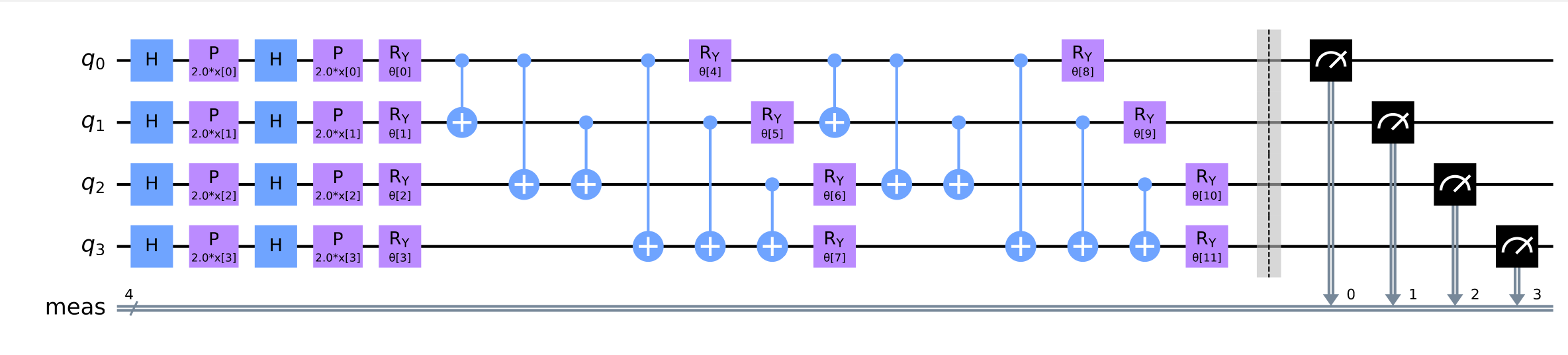}
    \caption{Complete quantum circuit for QNN-1.}
    \label{fig:qnn1_circ}
\end{figure}

\begin{figure}[h!]
    \centering
    \includegraphics[width=0.65\linewidth]{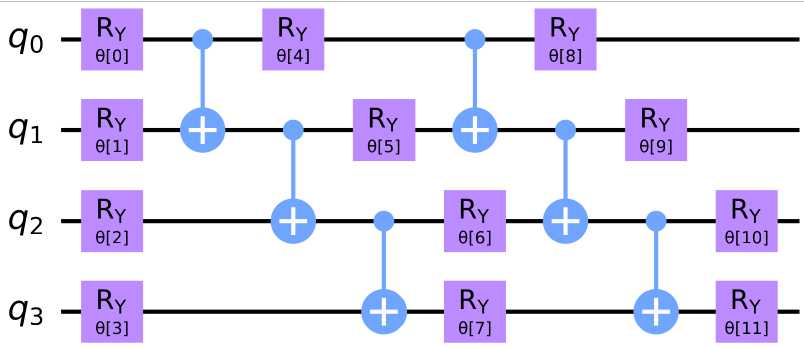}
    \caption{Ansatz with linear entanglement.}
    \label{fig:linear_ansatz_circ}
\end{figure}

\begin{figure}[h!]
    \centering
    \includegraphics[width=0.85\linewidth]{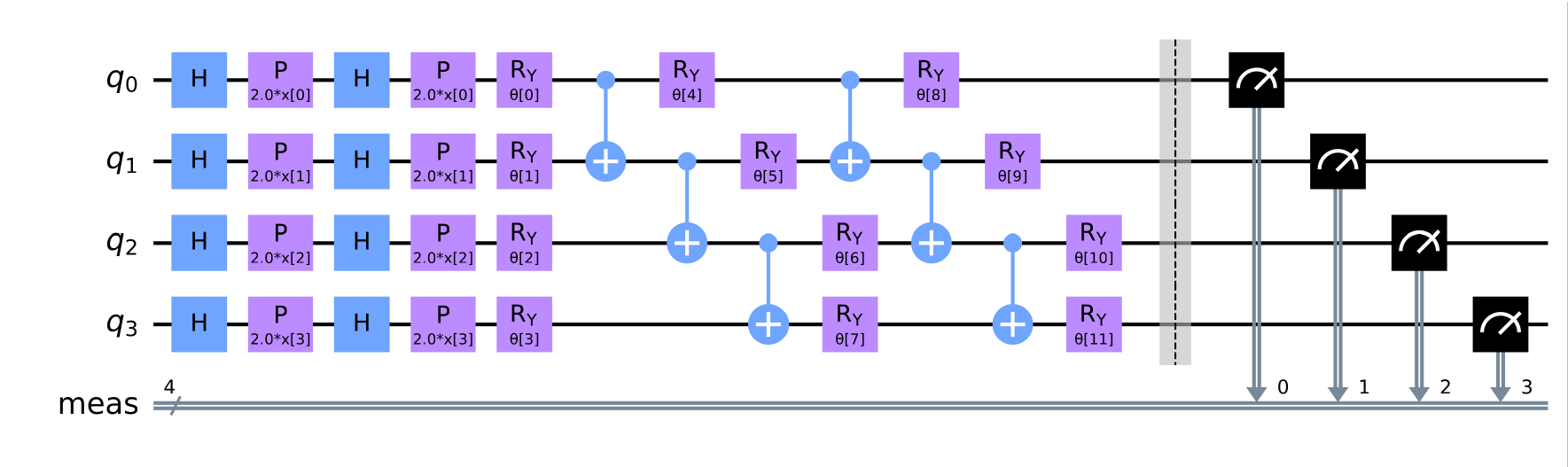}
    \caption{Complete quantum circuit for QNN-2.}
    \label{fig:qnn2_circ}
\end{figure}

\begin{figure}[h!]
    \centering
    \includegraphics[width=0.65\linewidth]{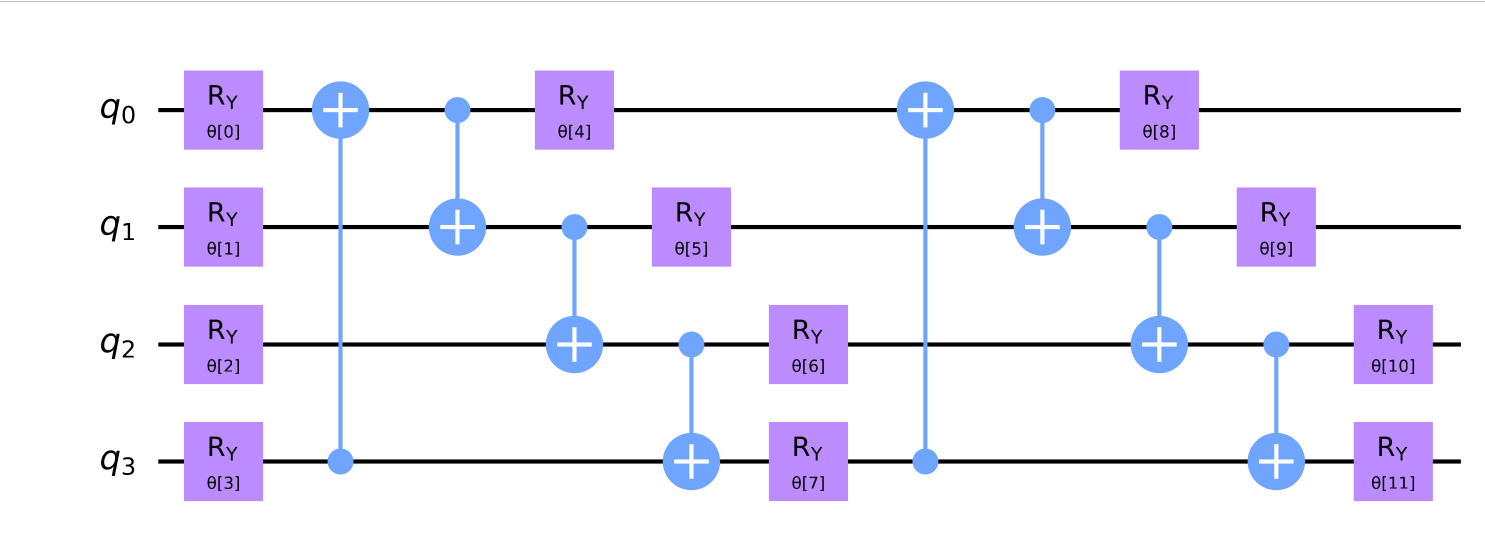}
    \caption{Ansatz with circular entanglement.}
    \label{fig:circular_ansatz_circ}
\end{figure}

\begin{figure}[h!]
    \centering
    \includegraphics[width=0.85\linewidth]{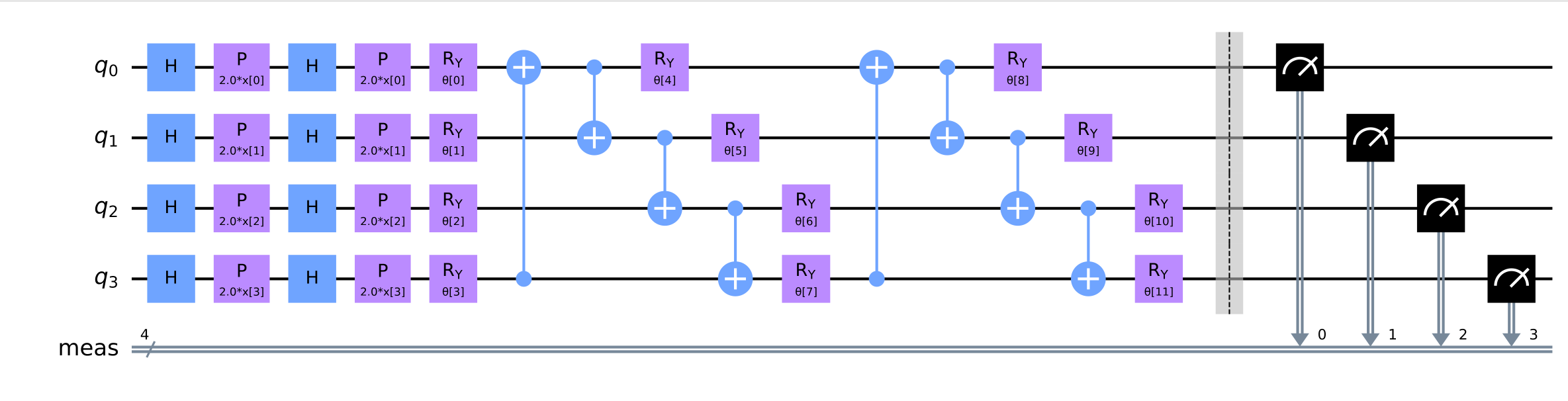}
    \caption{Complete quantum circuit for QNN-3.}
    \label{fig:qnn3_circ}
\end{figure}

\begin{figure}[h!]
    \centering
    \includegraphics[width=0.65\linewidth]{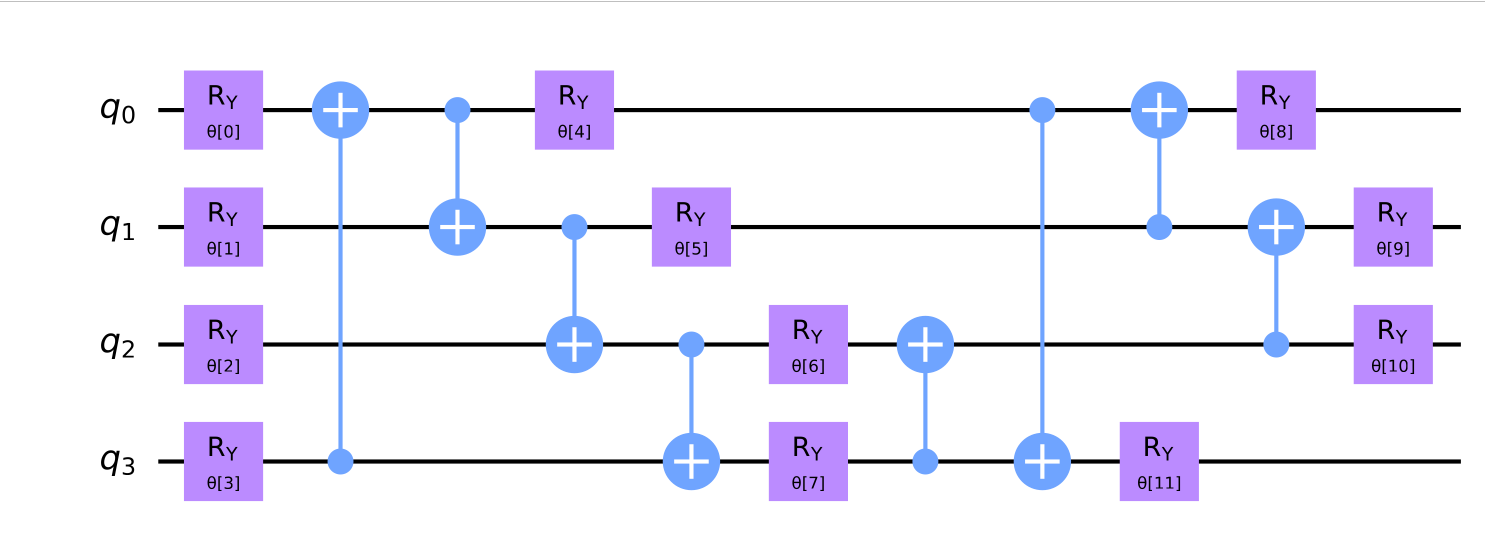}
    \caption{Ansatz with Shifted-Circular-Alternating (SCA) entanglement.}
    \label{fig:sca_ansatz_circ}
\end{figure}

\begin{figure}[h!]
    \centering
    \includegraphics[width=0.85\linewidth]{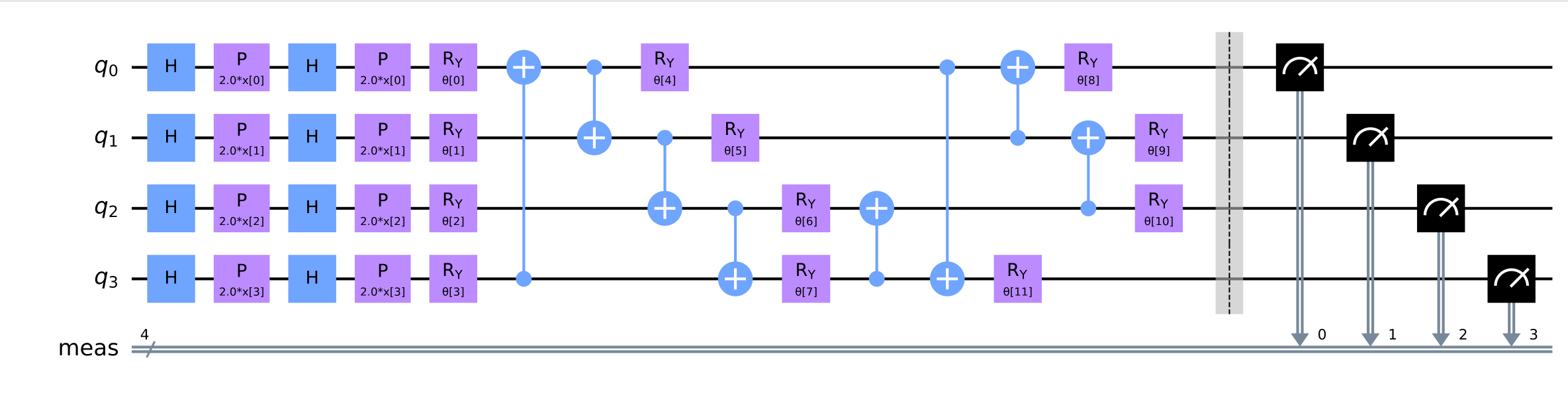}
    \caption{Complete quantum circuit for QNN-4.}
    \label{fig:qnn4_circ}
\end{figure}

\begin{figure}[h!]
    \centering
    \includegraphics[width=0.65\linewidth]{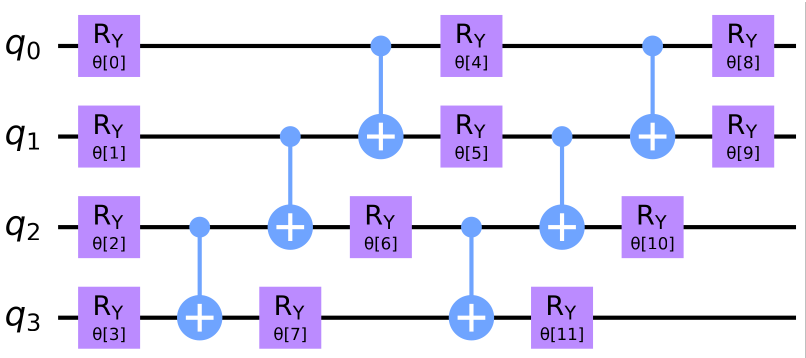}
    \caption{Ansatz with reverse linear entanglement.}
    \label{fig:reverse_linear_ansatz_circ}
\end{figure}

\begin{figure}[h!]
    \centering
    \includegraphics[width=0.85\linewidth]{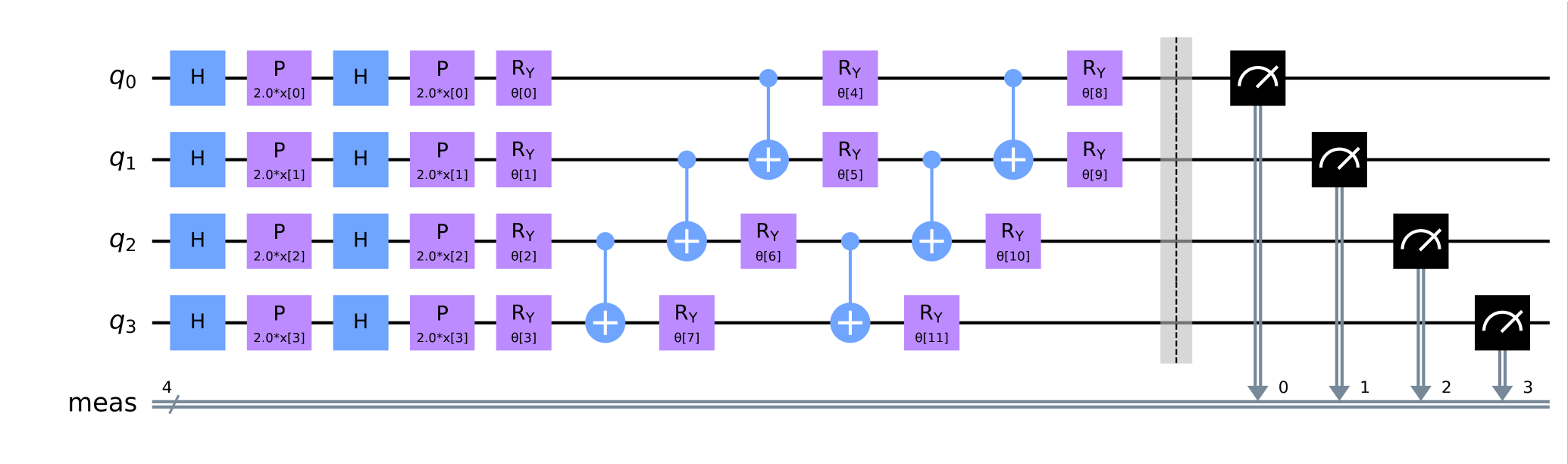}
    \caption{Complete quantum circuit for QNN-5.}
    \label{fig:qnn5_circ}
\end{figure}

\begin{figure}[h!]
    \centering
    \includegraphics[width=0.65\linewidth]{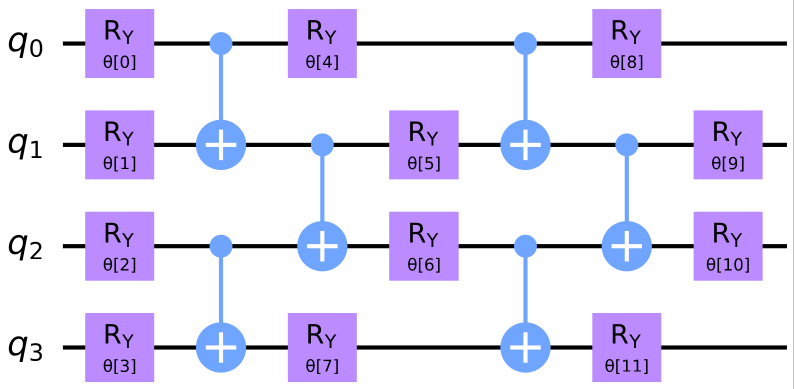}
    \caption{Ansatz with pairwise entanglement.}
    \label{fig:pairwise_ansatz_circ}
\end{figure}
\vspace*{-30pt}
\begin{figure}[h!]
    \centering
    \includegraphics[width=0.85\linewidth]{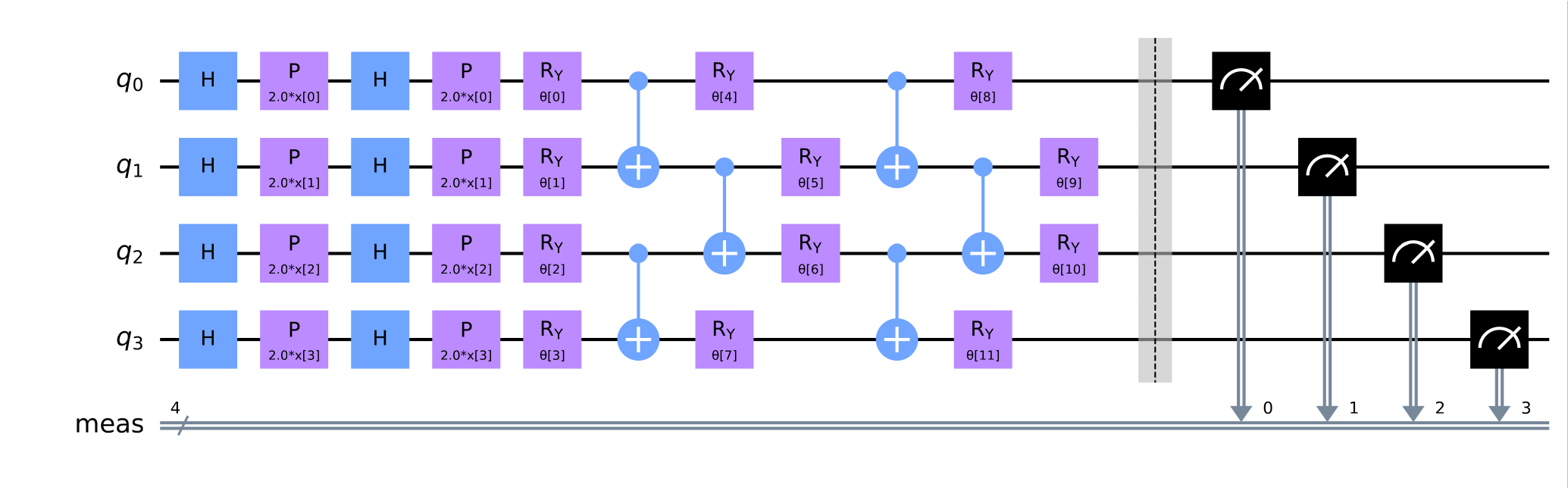}
    \caption{Complete quantum circuit for QNN-6.}
    \label{fig:qnn6_circ}
\end{figure}

\clearpage
\section{Actual vs. Predicted scatter plots for QNN models across the test set, corresponding to the $R^2$ and $RMSE$ values in Table~\ref{tab:comparison_qnn_vs_ml_table}}\label{append2}

\begin{figure}[h!]
    \centering
    \includegraphics[width=1.1\linewidth]{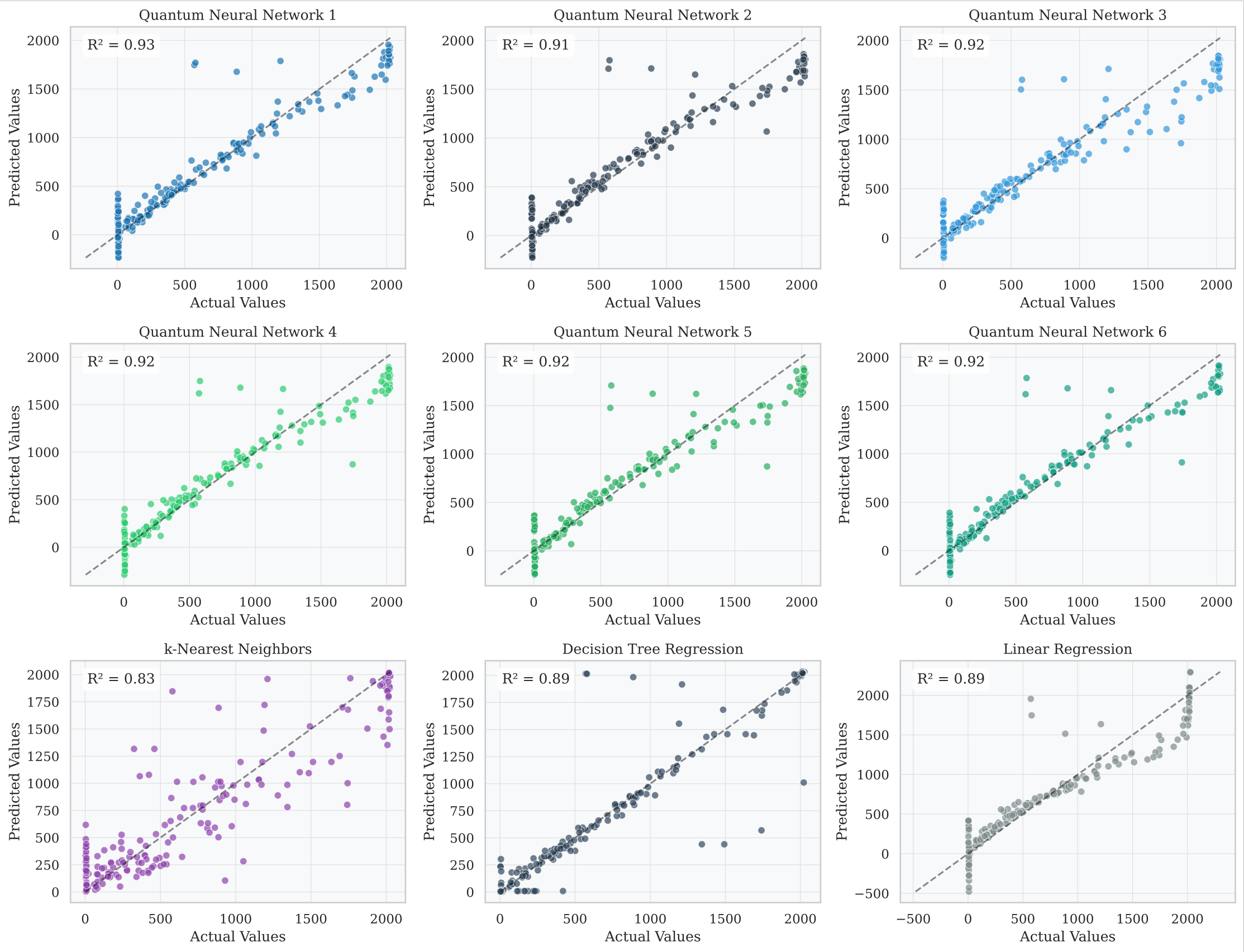}
    \caption{Scatter plot of actual vs predicted values comparing QNN and classical models for a test set of 200 samples}
    \label{fig:act_vs_pred_scatter_200}
\end{figure}

\begin{figure}[h!]
    \centering
    \includegraphics[width=1.1\linewidth]{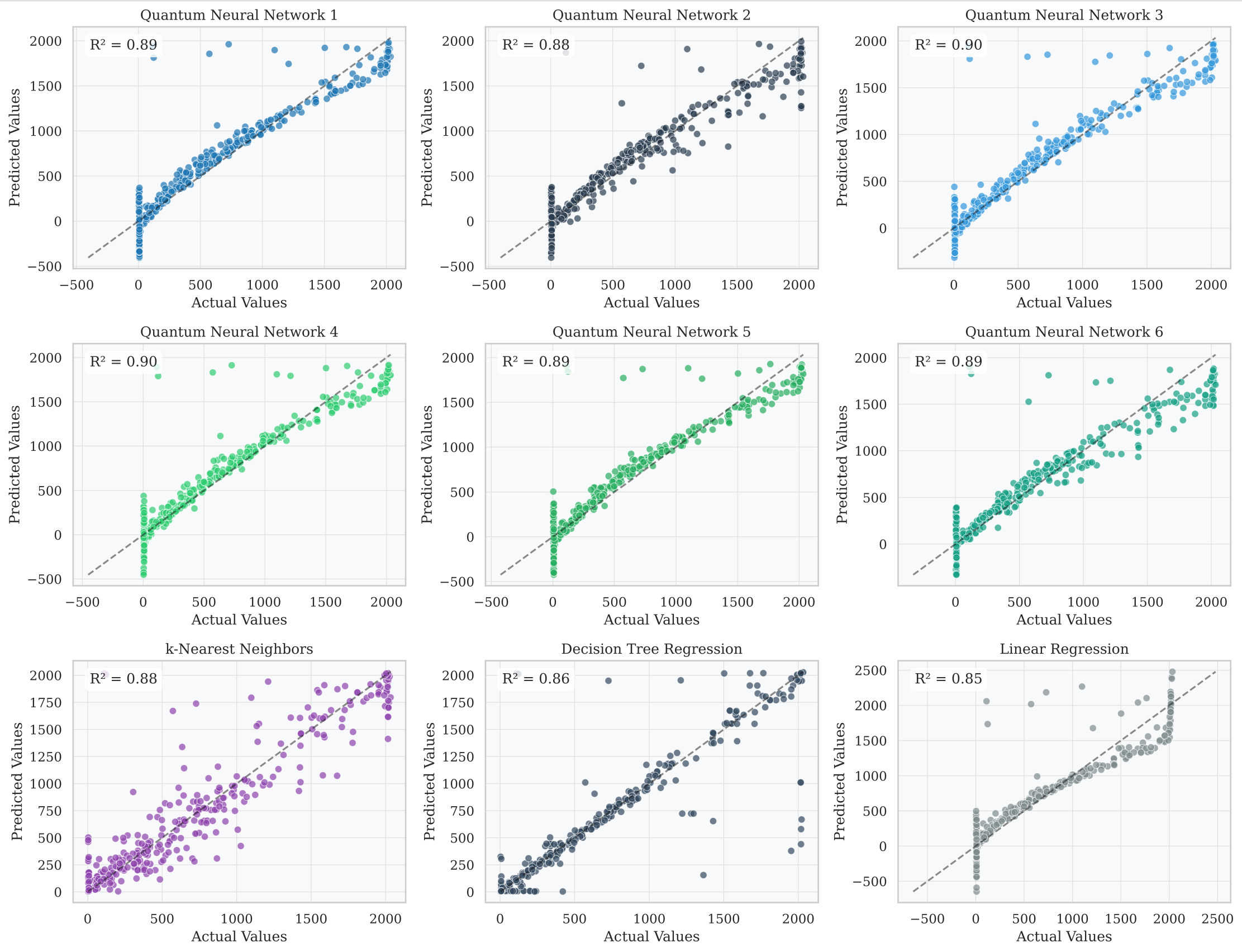}
    \caption{Scatter plot of actual vs predicted values comparing QNN and classical models for a test set of 400 samples}
    \label{fig:act_vs_pred_scatter_400}
\end{figure}

\begin{figure}[h!]
    \centering
    \includegraphics[width=1.1\linewidth]{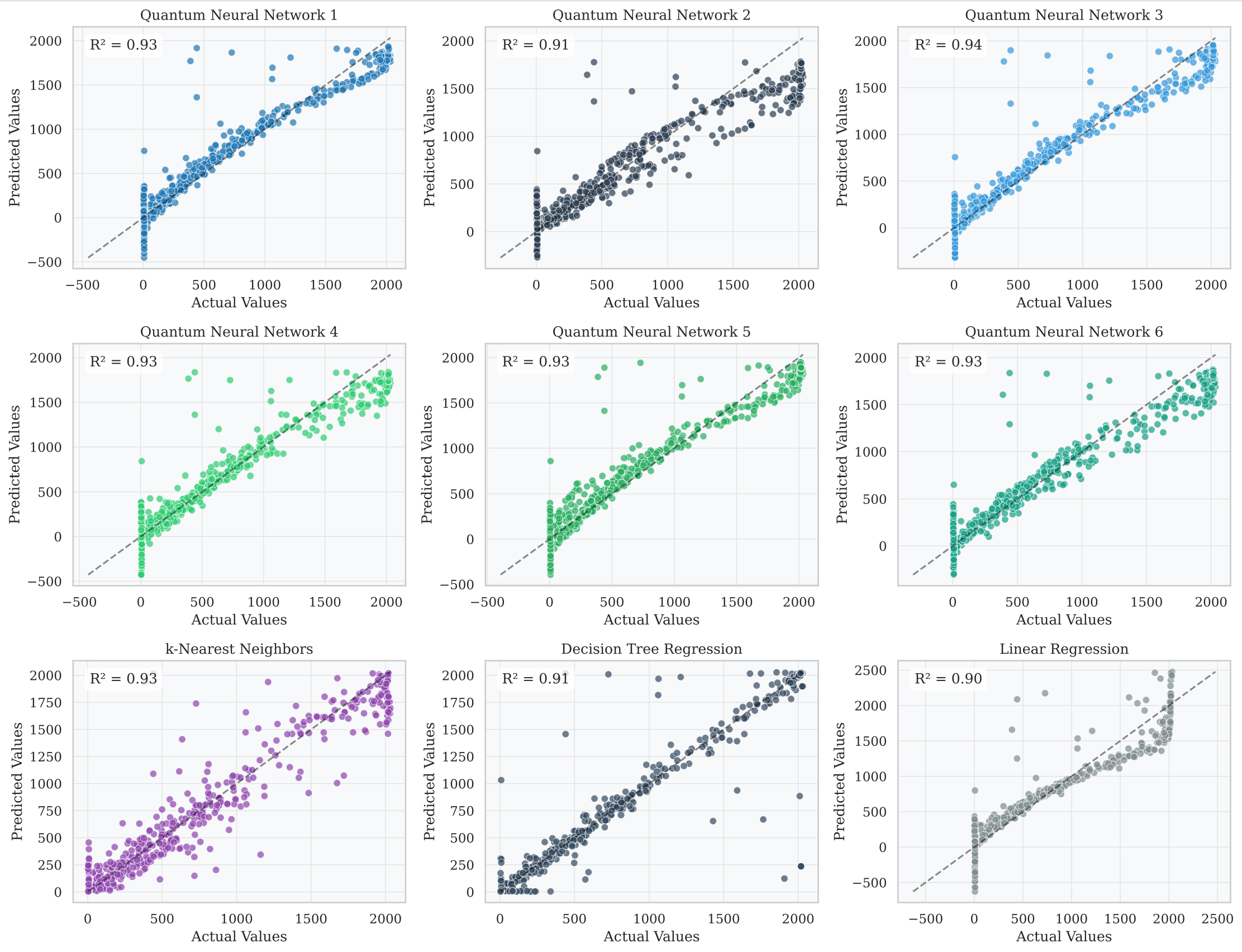}
    \caption{Scatter plot of actual vs predicted values comparing QNN and classical models for a test set of 600 samples}
    \label{fig:act_vs_pred_scatter_600}
\end{figure}

\begin{figure}[h!]
    \centering
    \includegraphics[width=1.1\linewidth]{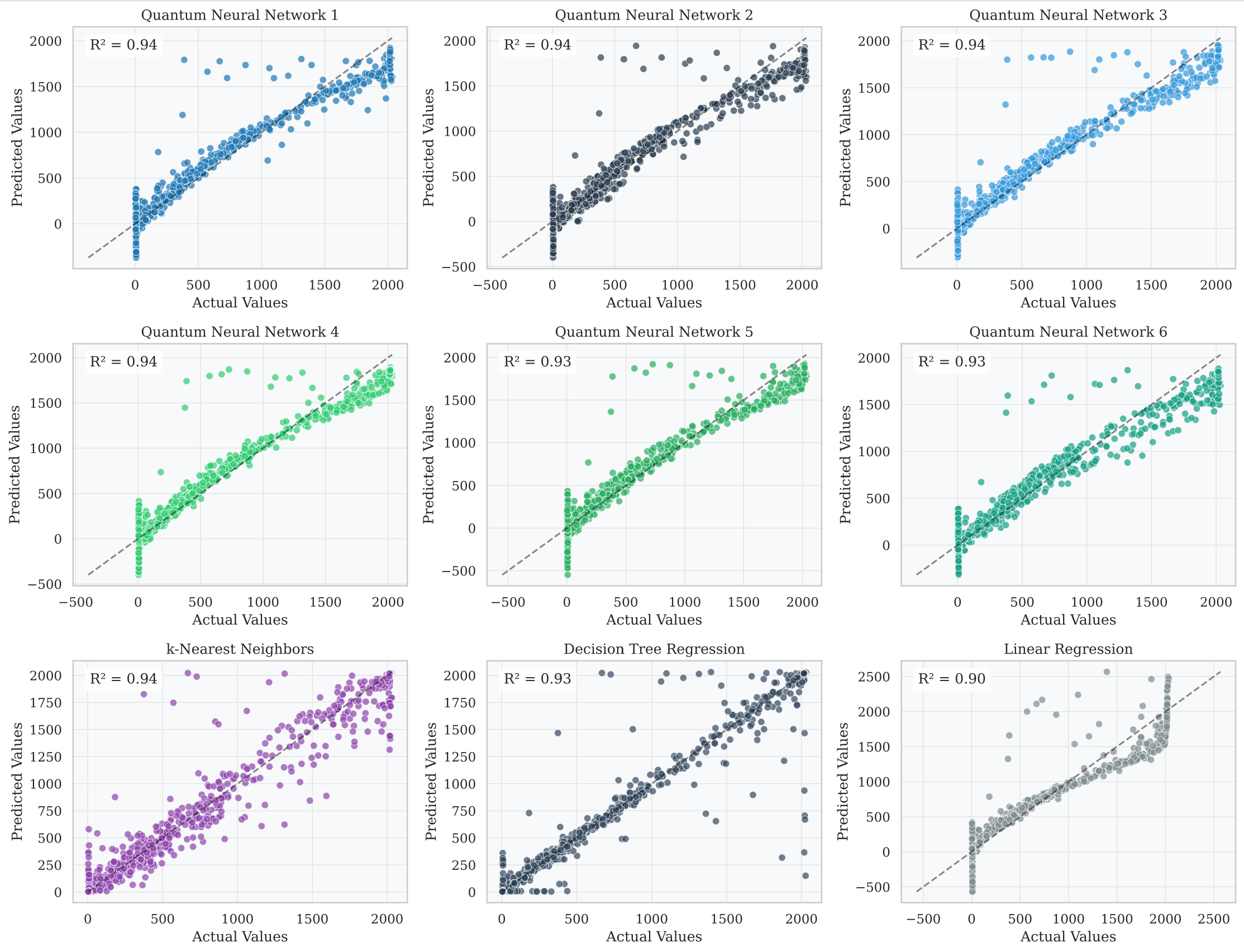}
    \caption{Scatter plot of actual vs predicted values comparing QNN and classical models for a test set of 800 samples}
    \label{fig:act_vs_pred_scatter_800}
\end{figure}

\clearpage
\section{Predicted and actual value comparison over data index for each model}\label{append3}

\begin{figure}[h!]
    \centering
    \includegraphics[width=1.1\linewidth]{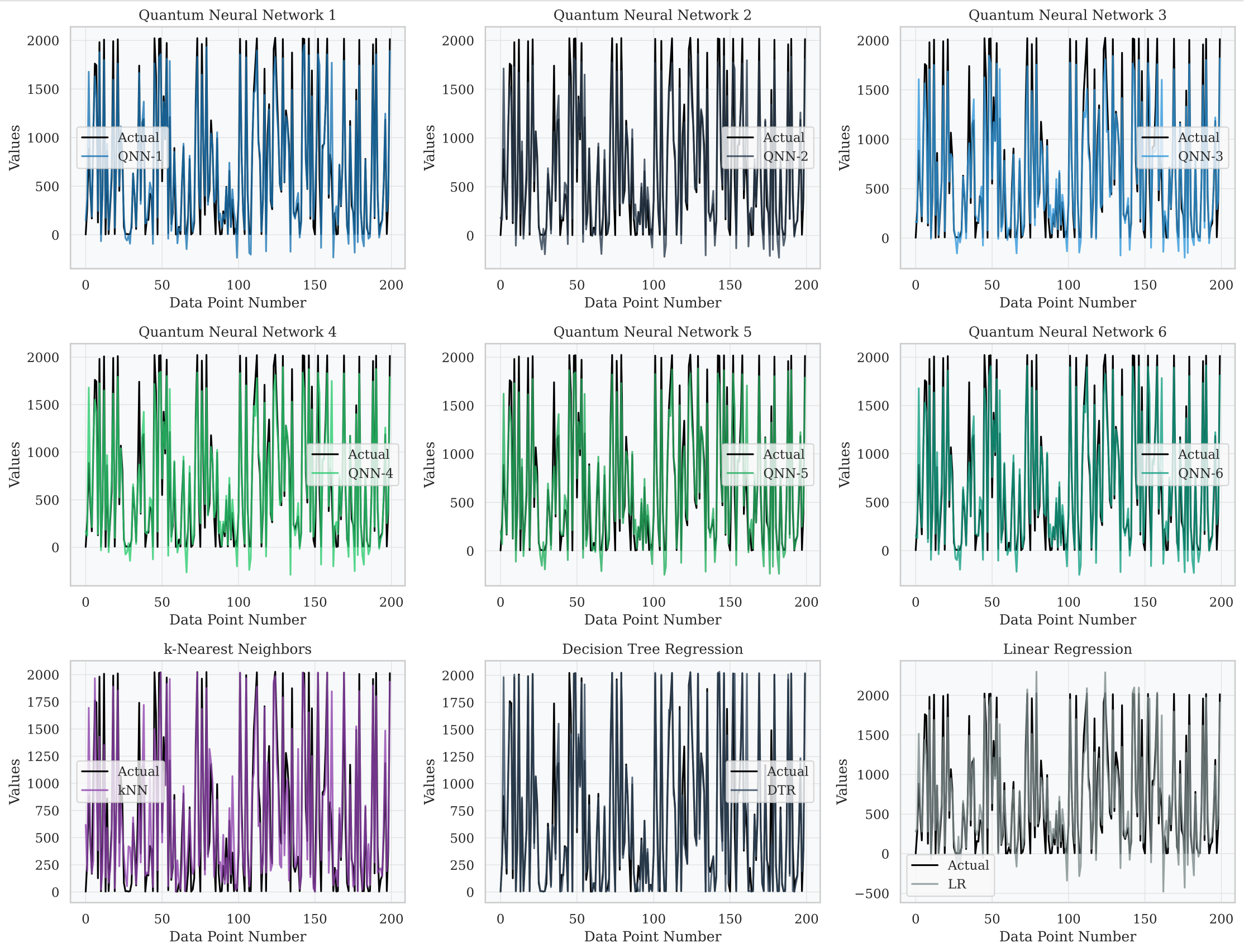}
    \caption{Line plot of actual vs predicted values comparing QNN and classical models for a test set of 200 samples}
    \label{fig:act_vs_pred_overlay_200}
\end{figure}

\begin{figure}[h!]
    \centering
    \includegraphics[width=1.1\linewidth]{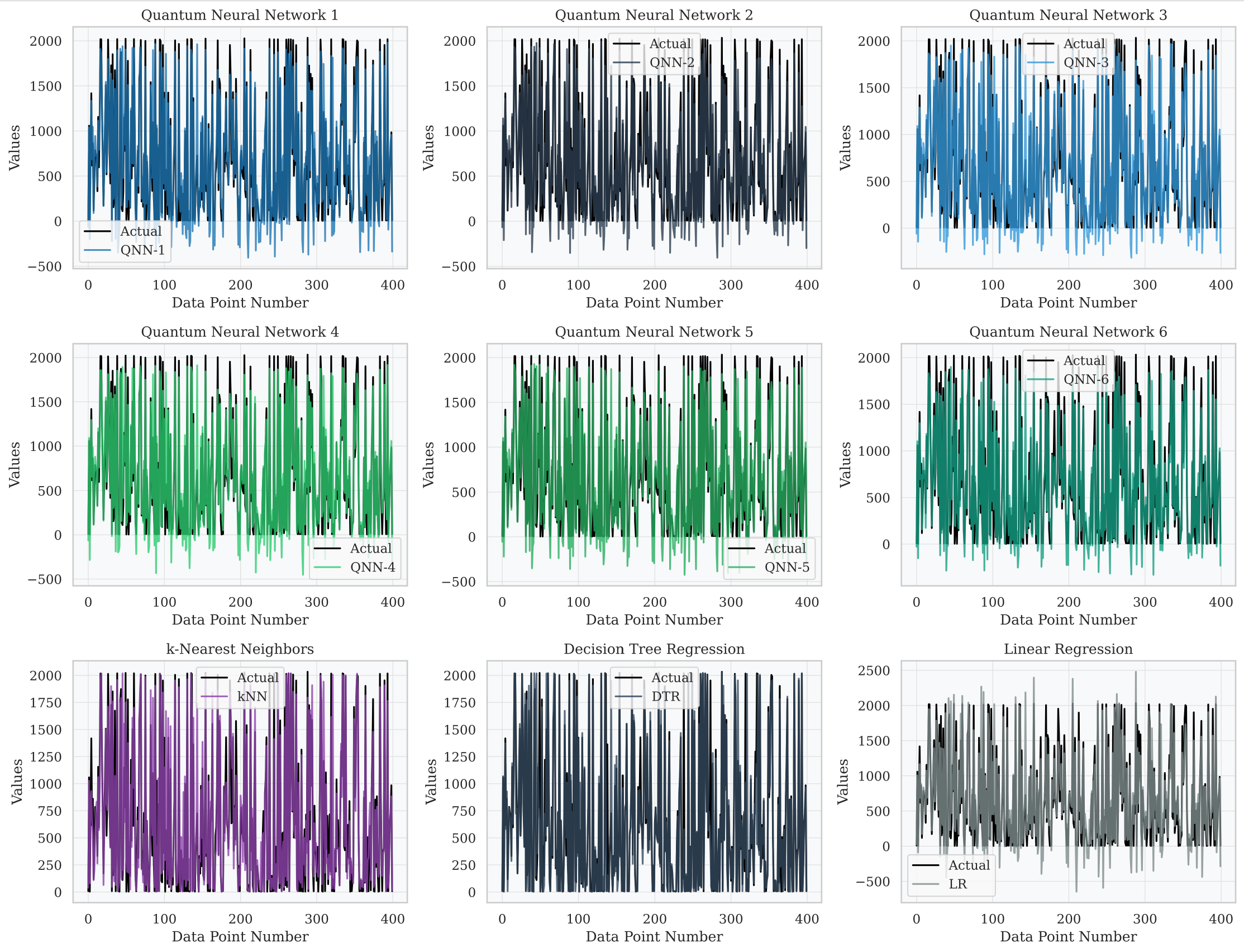}
    \caption{Line plot of actual vs predicted values comparing QNN and classical models for a test set of 400 samples}
    \label{fig:act_vs_pred_overlay_400}
\end{figure}

\begin{figure}[h!]
    \centering
    \includegraphics[width=1.1\linewidth]{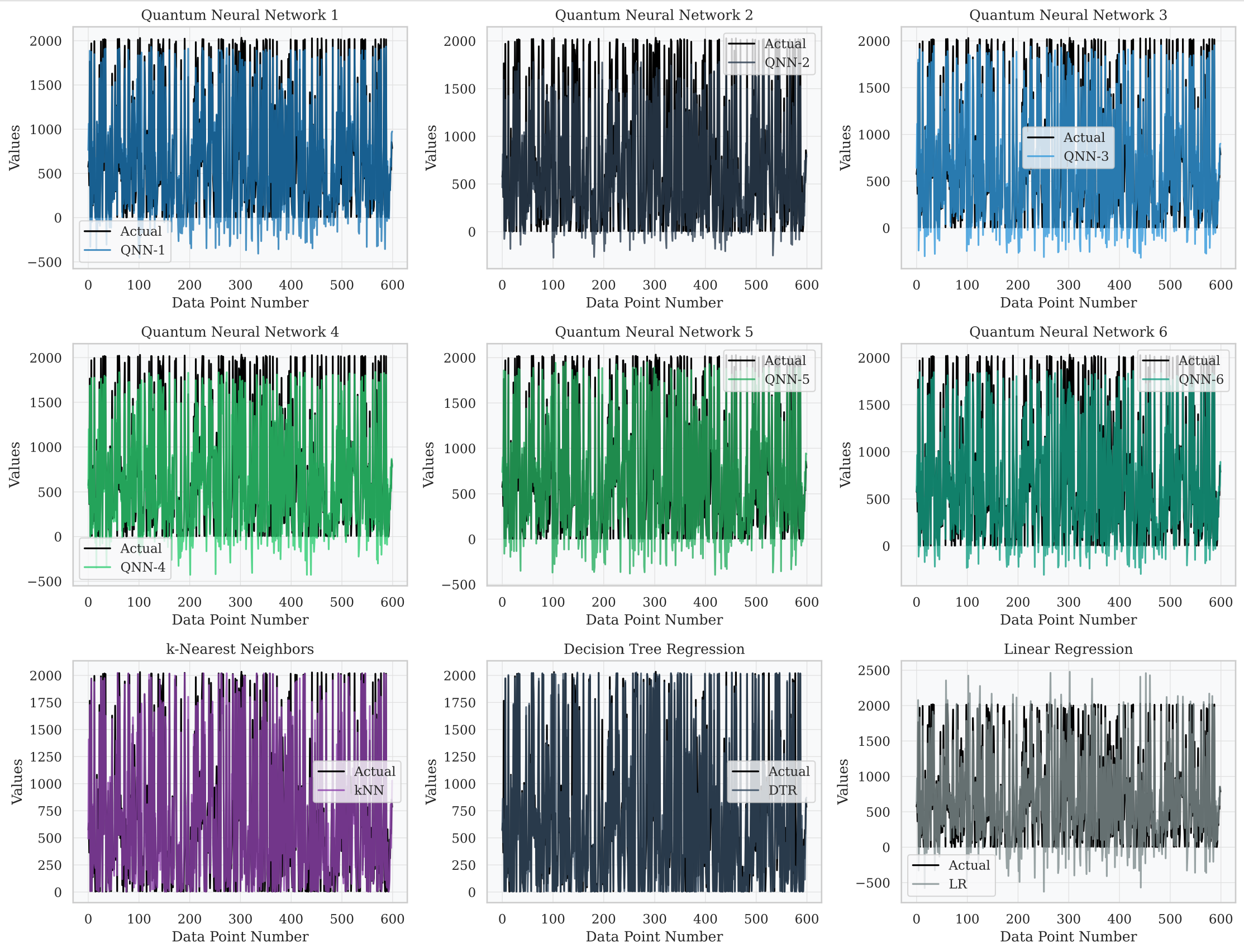}
    \caption{Line plot of actual vs predicted values comparing QNN and classical models for a test set of 600 samples}
    \label{fig:act_vs_pred_overlay_600}
\end{figure}

\begin{figure}[h!]
    \centering
    \includegraphics[width=1.1\linewidth]{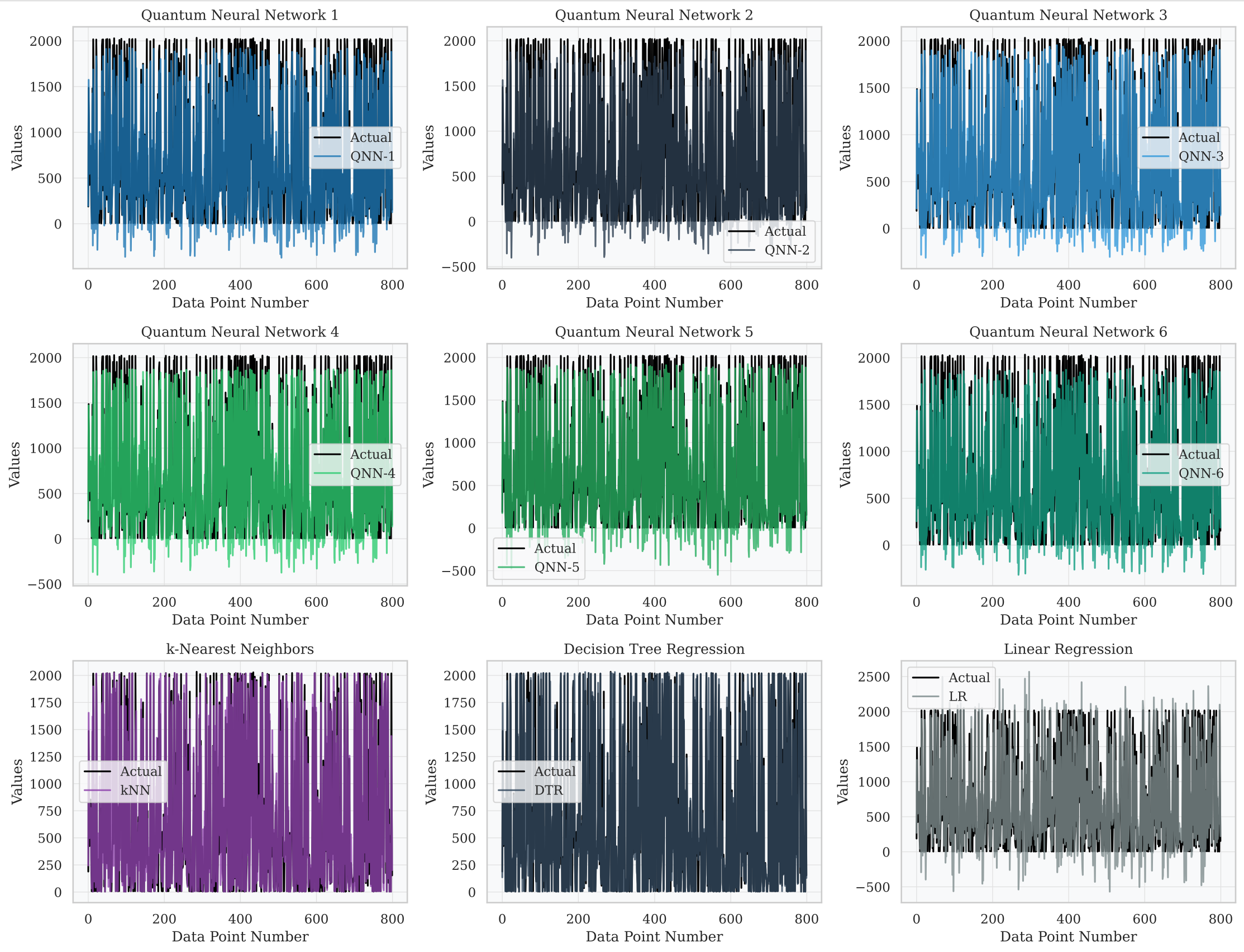}
    \caption{Line plot of actual vs predicted values comparing QNN and classical models for a test set of 800 samples}
    \label{fig:act_vs_pred_overlay_800}
\end{figure}

\end{document}